\documentclass [10pt]{article}
\usepackage[utf8]{inputenc}
\usepackage{amsmath}
\usepackage{amsfonts}
\usepackage{amssymb}
\usepackage{graphicx}
\usepackage{subcaption}
\usepackage{soul}
\usepackage{parskip}
\usepackage{natbib}
\usepackage{bbm}
\usepackage{algorithm}
\usepackage{bm}
\usepackage[noend]{algpseudocode}
\usepackage{booktabs}
\usepackage{tabularx}
\usepackage{xcolor}
\usepackage{url}
\usepackage{algorithmicx}
\usepackage{multirow}
\usepackage{lscape}
\usepackage{pifont}
\usepackage{CJKutf8}

\title{\textbf{Estimating the Error of Large Language Models at \\
Pairwise Text Comparison}}

\author{\textsc{Tianyi Li}\thanks{tianyi.li@cuhk.edu.hk} \\[1ex] 
Department of Decisions, Operations and Technology, CUHK \\
}
\date{}

\begin{document}

\maketitle

\begin{abstract}

\noindent We measure LLMs' output error at pairwise text comparison, noting the probability of error in their preferences. Our method does not rely on the ground truth and supports two scenarios: (i) uniform error rate regardless of the order of comparison, estimated with two comparisons for each text pair with either text placed first; (ii) binary positional bias assuming distinct error rates for the two orders of comparison, estimated with repeated comparisons between the texts. The Copeland counting constructs a ranking over the compared texts from pairwise preferences; the ranking reveals the poor scalability of LLM-based pairwise comparison and helps yield the estimates for LLMs' error rates. We apply the method to six LLMs (ChatGPT, Claude, DeepSeek, Gemini, Grok, Qwen) with five types of text input and obtain consistent estimates of LLMs' error. In general, the measured two positional bias terms are similar, close to the uniform error. Considering both the error rates and the robustness to the variation of prompts, Claude obtained the most desirable performance in this experiment. Our model outperforms the biased Bradley-Terry model and the commutativity score in indicating LLMs' error at this task. 
\end{abstract}

Hallucination \citep{JLet2023,CDet2024,COet2025,HYet2025}, which leads to deceptive behavior \citep{H2024}, is a rooted problem for the output from LLMs, where factual errors \citep{ABet2024} contaminate the output due to generative models' stochastic nature. The level of hallucination can be quantified by comparing the LLM output to the ground truth \citep[e.g.,][]{SCet2023}. When the ground truth is not available \citep[e.g.,][]{WYet2024}, particular settings could be conceived to help reveal LLMs' errors. For example, ask the LLM the same question multiple times and see if the answers are consistent \citep[e.g.,][]{MLet2023}. 

The output's consistency can be evaluated in a rigorous way when the answers are indicative rather than descriptive. A good setting in this regard is to use the LLM to compare input objects and ask it to output a preference. The delivered instances of preference from an ensemble of inquiries then help quantify the LLM output's rate of error. Further, based on the results of comparison, a ranking over the compared objects can be constructed, which can be contrasted with the ground-truth ranking if there exists one \citep[e.g.,][]{LZet2024}.

The basic format for this task is pairwise text comparison \citep[e.g.,][]{CWet2023,SCet2023,LGet2024}, applicable to LLMs of different types and performance levels while extensible to, for example, setwise comparison \citep{ZZet2024} and other types of input such as at multimodel LLMs \citep[e.g.,][]{KFet2023}. A key aspect of evaluating LLMs' output error via pairwise comparison is to swap the order of comparison and check the consistency of the two preferences, which is often violated -- the indication of preference (i.e., object 1 or 2) remains unchanged when the comparison order is swapped. The hallucination of the LLM output exhibits here as order inconsistency \citep{ZTet2024}, sometimes termed as positional bias \citep{WLet2023,ZCet2023}: LLMs' propensity to favor certain positional configurations of the input over others. As noted in \citep{ZCet2023}, positional bias at texts is prevalent in human decision-making \citep{B1984} and other machine-learning decision settings, e.g., learning-to-rank from click data \citep{WLet2018} or question-answering models \citep{KLet2020}. It is important to study such a fallacy in this new context of human-machine exchange \citep{PRet2025}.

Another issue with employing LLMs in pairwise comparison is its poor scalability: the number of comparisons grows quadratically as the number of objects increases \citep{ZCet2023,LRet2024}, and thus constructing a ranking from comparison results requires sub-sampling and algorithmic caveats to speed up \citep[e.g.,][]{QJet2023}. Besides the difficulty in computation, one can imagine that the reliability of objects' ranking based on pairwise comparison also deteriorates when it scales up. It is useful to validate this conjecture with formal constructs.

In this work, we propose a method to characterize LLMs' error-prone output when recruiting them for pairwise text comparison. The method captures the probability of error in the LLM-output preferences, considering two scenarios: (i) a uniform error rate regardless of the order of comparison, estimated with two comparisons for each text pair with either text placed first; (ii) the positional bias which assumes distinct error rates for the two orders of comparison, estimated with multiple instances of preference collected from a series of comparisons between the two objects. The classic Copeland counting \citep{C1951,SW2018} is applied to construct a ranking over the compared objects from pairwise preferences. The obtained Copeland ranking demonstrates the poor scalability of LLM-based pairwise comparison and helps yield the estimates for LLMs' error rates in this ground-truth-free setting. We apply the method to data obtained from six popular LLMs (ChatGPT, Claude, DeepSeek, Gemini, Grok, Qwen) with five types of text input (pseudo-word paragraphs, pseudo paragraphs, advertising slogans, short poems, academic abstracts); results yield consistent estimates of LLMs' error across the two scenarios. Our estimates align with the commutativity scores \citep{LGet2024} while indicating LLMs' error to a better extent. We construct a biased Bradley-Terry (BT) model \citep{BT1952} for alternative error estimation; our model outperforms the biased BT model in delivering coherent estimates.

\section*{Methods}

Consider an ensemble of $N$ texts. Suppose the ground-truth outcome of comparing text $i$ to text $j$ is $y_{ij}\in [-1,1]$, and the LLM-output preference is $\hat{y}_{ij}\in [-1,1]$; for both, -1/1 means text $i$ is not preferred/preferred to $j$. When swapping the order of comparison, the ground-truth/LLM-output preference is $y_{ji}$ and $\hat{y}_{ji}$, respectively. For the ground truth, $y_{ij} = - y_{ji}$, while it is likely that $\hat{y}_{ij} = \hat{y}_{ji}$, as the LLM output may contain error. Denote the matrices $\mathbf{Y_g}=\{y_{ij}\}$ and $\mathbf{Y}=\{\hat{y}_{ij}\}$. The commutativity \citep[e.g.,][]{R2020} score $S_{com} \in [0, 1]$ proposed in \citep{LGet2024} evaluates the proportion of comparisons having the opposite preference when the comparison order is swapped, i.e., the proportion of cases demonstrating inconsistent output across the two instances of comparison. It can be calculated from the matrix $\mathbf{Y}$: $S_{com} = 2/N(N-1)\sum_{i<j} \mathbbm{1} [\hat{y}_{ij} = \hat{y}_{ji}]$. A small/large $S_{com}$ indicates a small/large error rate over the ensemble of comparisons. 

\subsection*{Uniform error} 

Suppose that the LLM-output preference $\hat{y}_{ij}$ contains error with probability $\epsilon$, uniformly for any $i,j$:
\begin{equation}
\forall i,j,\ \ \ \hat{y}_{ij} = \left\{
\begin{aligned}
&y_{ij} \text{  [True]}, \ \ P = 1-\epsilon, \\
-&y_{ij}\text{  [False]}, \ \ P = \epsilon. \\
\end{aligned}
\right.
\end{equation}
Averaging the outcomes $\hat{y}_{ij}$ and $\hat{y}_{ji}$ from the two instances of comparison between $i$ and $j$, we obtain
\begin{equation}
\hat{z}_{ij} = \frac{\hat{y}_{ij}+(-\hat{y}_{ji})}{2},
\end{equation}
which constitutes the matrix $\mathbf{Z}=\{\hat{z}_{ij}\}$. $\mathbf{Z}$ is diagonally antisymmetric: $\hat{z}_{ij} = -\hat{z}_{ji}$. There are four cases for the outcome at $\hat{z}_{ij}$: (i) when $\hat{y}_{ij}$ is False and $\hat{y}_{ji}$ is True, $\hat{z}_{ij} = 0$; (ii) when $\hat{y}_{ij}$ is True and $\hat{y}_{ji}$ is True, $\hat{z}_{ij} = y_{ij}$; (iii) when $\hat{y}_{ij}$ is False and $\hat{y}_{ji}$ is False, $\hat{z}_{ij} = -y_{ij}$; (iv) when $\hat{y}_{ij}$ is True and $\hat{y}_{ji}$ is False, $\hat{z}_{ij} = 0$. Cases (i) and (iv) yield the same outcome; the probabilities for different values of $\hat{z}_{ij}$ are
\begin{equation}
\left\{
\begin{aligned}
&P(\hat{z}_{ij} = 0) &&= 2\epsilon(1-\epsilon), \\
&P(\hat{z}_{ij} = y_{ij}\ \text{[True outcome (TO)]}) &&= (1-\epsilon)^2, \\
&P(\hat{z}_{ij} = -y_{ij}\ \text{[Inverse outcome (IO)]}) &&= \epsilon^2. \\
\end{aligned}
\right.
\label{eq-p}
\end{equation}
We use the Copeland counting to construct a ranking of the texts. The Copeland ranking relies on the scores that sum on the matrix $\mathbf{Z}$: the score for text $i$ is 
\begin{equation}
S_{Copeland}^Z(i) = \sum_j^{N-1} \hat{z}_{ij},
\end{equation}
We analyze the probability for a text to have the correct score $S_{Copeland}$ from the biased LLM-preferences.

First, consider the text having the ground-truth ranking $\#1$. This text $R_1$ wins all comparisons: $y_{R_1j} = 1$ for any $j\neq R_1$, and thus $S_{Copeland}^Z(R_1) = N-1$. To realize this score $S_{Copeland}^Z(R_1)$, all elements in the sum need to be true, which has the probability
\begin{equation}
P(R_1,\text{TO}) = \prod_{N-1}P(\hat{z}_{R_1j} = y_{R_1j}\ \text{[True outcome (TO)]}) = (1-\epsilon)^{2(N-1)}.
\end{equation}
Next, consider the text having the ground-truth ranking $\#2$. This text $R_2$ wins all comparisons except the comparison with text $R_1$: $y_{R_2R_1} = -1$, and $y_{R_2j} = 1$ for any $j\neq R_1, j\neq R_2$, and thus $S_{Copeland}^Z(R_1) = N-3$. To realize this score $S_{Copeland}^Z(R_2)$, either (i) all elements in the sum are true, or (ii) $\hat{z}_{R_2R_1}$ and another $\hat{z}_{R_2j}$ for an arbitrary $j$ are false, and the rest $N-3$ elements are true. This has the probability
\begin{equation}
P(R_2,\text{TO}) = (1-\epsilon)^{2(N-1)} + (1-\epsilon)^{2(N-3)}C_1^1C_{N-2}^1\epsilon^{2\times2}.
\end{equation}
Similarly, for $S_{Copeland}^Z(R_3)$ to be true, there are three possible cases, and their combined probability is 
\begin{equation}
P(R_3,\text{TO}) = (1-\epsilon)^{2(N-1)} + (1-\epsilon)^{2(N-3)}C_2^1C_{N-3}^1\epsilon^{2\times2} + (1-\epsilon)^{2(N-5)}C_2^2C_{N-3}^2\epsilon^{2\times4}.
\end{equation}
For $S_{Copeland}^Z(R_m)$ to be true, there are $m$ possible cases, and the combined probability is 
\begin{equation}
\begin{aligned}
P(R_m,\text{TO}) = (1-\epsilon)^{2(N-1)} + (1-\epsilon)^{2(N-3)}C_{m-1}^1C_{N-m}^1\epsilon^{2\times2} + (1-\epsilon)^{2(N-5)}C_{m-1}^2C_{N-m}^2\epsilon^{2\times4} + ... \\
+ (1-\epsilon)^{2[N-1-2(m-1)]}C_{m-1}^{m-1}C_{N-m}^{m-1}\epsilon^{2\times2(m-1)}. 
\end{aligned}
\end{equation}
This ends halfway at the rank: $N-m \geq m-1 \Rightarrow m \leq (N+1)/2$. For the second half of the rank, $m > (N+1)/2$, the probabilities mirror the first half, with the substitution $m-1 \rightarrow N-m$. There is
\begin{equation}
\begin{aligned}
P(R_m,\text{TO}) = (1-\epsilon)^{2(N-1)} + (1-\epsilon)^{2(N-3)}C_{N-m}^1C_{m-1}^1\epsilon^{2\times2} + (1-\epsilon)^{2(N-5)}C_{N-m}^2C_{m-1}^2\epsilon^{2\times4} + ... \\
+ (1-\epsilon)^{2[N-1-2(N-m)]}C_{N-m}^{N-m}C_{m-1}^{N-m}\epsilon^{2\times2(N-m)};
\end{aligned}
\end{equation}
for example, notably,
\begin{equation}
\begin{aligned}
P(R_{N-1},\text{TO}) &= P(R_2,\text{TO}) = (1-\epsilon)^{2(N-1)} + (1-\epsilon)^{2(N-3)}C_1^1C_{N-2}^1\epsilon^{2\times2}, \\
P(R_{N},\text{TO}) &= P(R_1,\text{TO}) = (1-\epsilon)^{2(N-1)}.
\end{aligned}
\end{equation}
To investigate the scalability of this Copeland ranking based on pairwise comparisons, we analyze the change in the probability of obtaining the correct $S_{Copeland}^Z(R_{\bullet})$, $P(R_{\bullet},\text{TO})$, as $N$ goes up. For the $\#1$ text $R_1$, 
\begin{equation}
\frac{dP(R_1,\text{TO})}{dN} = \frac{d(1-\epsilon)^{2(N-1)}}{dN} < 0,
\end{equation}
i.e., for all values of $\epsilon$, $P(R_1,\text{TO})$ always decreases as $N$ goes up. 

For the $\#2$ text $R_2$, 
\begin{equation}
\begin{aligned}
\frac{dP(R_2,\text{TO})}{dN} &= \frac{d[(1-\epsilon)^{2(N-1)} + (1-\epsilon)^{2(N-3)}C_1^1C_{N-2}^1\epsilon^{2\times2}]}{dN} \\
&= (1-\epsilon)^{2(N-3)}\{\epsilon^4 + 2ln(1-\epsilon)[(1-\epsilon)^4+(N-2)\epsilon^4]\}. \\
\end{aligned}
\end{equation}
Then
\begin{equation}
\begin{aligned}
\frac{dP(R_2,\text{TO})}{dN} > 0 \Rightarrow & \epsilon^4 + 2ln(1-\epsilon)[(1-\epsilon)^4+(N-2)\epsilon^4] > 0 \\
\Rightarrow & N - 2 < -\frac{1}{2ln(1-\epsilon)} - (\frac{1-\epsilon}{\epsilon})^4 < 0, \ \ \text{for $\epsilon \in [0,1)$}.
\end{aligned}
\end{equation}
This cannot be satisfied with positive $N$; thus $dP(R_2,\text{TO})/dN < 0$ uniformly for all $\epsilon$, same as the $R_1$ case.

Similar derivations can establish that $dP(R_m,\text{TO})/dN < 0$ for all $\epsilon$ and $m$, which can be visualized with simulations (Figure 1). $P(R_{\bullet},TO)$ approaches $P(R_{1},TO)$ when $\epsilon$ is small, as the terms with the $\epsilon$ multipliers vanish; the difference between $P(R_{1},TO)$ and $P(R_{2},TO)$ can be seen, for example, at $\epsilon=0.5$. Overall, this concludes that for a text of any ground-truth ranking, the probability of obtaining its correct score $S_{Copeland}$ decreases uniformly as $N$ increases, i.e., the Copeland ranking constructed from $\mathbf{Z}$ is NOT scalable.

\subsection*{Positional bias} 

The positional bias considers that the error rates are different when the text is in the first or second place of the comparison: when the better text is placed first/second, the error rate is $\epsilon_+/\epsilon_-$. This extends the uniform error probability to:
\begin{equation}
\begin{aligned}
\ \ \ \ \ \ \ \ y_{ij} = 1, \ \ \ &&\hat{y}_{ij} = \left\{
\begin{aligned}
&y_{ij} \text{  [True]}, \ \ P = 1-\epsilon_+, \\
-&y_{ij}\text{  [False]}, \ \ P = \epsilon_+. \\
\end{aligned}
\right. 
\text{ and} \ \ \ \hat{y}_{ji} = \left\{
\begin{aligned}
&y_{ji} \text{  [True]}, \ \ P = 1-\epsilon_-, \\
-&y_{ji}\text{  [False]}, \ \ P = \epsilon_-. \\
\end{aligned}
\right.  
\\
\ \ \ \ \ \ \ \ y_{ij} = -1, \ \ \ &&\hat{y}_{ij} = \left\{
\begin{aligned}
&y_{ij} \text{  [True]}, \ \ P = 1-\epsilon_-, \\
-&y_{ij}\text{  [False]}, \ \ P = \epsilon_-. \\
\end{aligned}
\right. 
\text{ and} \ \ \ \hat{y}_{ji} = \left\{
\begin{aligned}
&y_{ji} \text{  [True]}, \ \ P = 1-\epsilon_+, \\
-&y_{ji}\text{  [False]}, \ \ P = \epsilon_+. \\
\end{aligned}
\right. 
\end{aligned}
\label{eq-pb}
\end{equation}
The probabilities for different values of $\hat{z}_{ij}$ are now
\begin{equation}
\left\{
\begin{aligned}
&P(\hat{z}_{ij} = 0) &&= \epsilon_+(1-\epsilon_-)+\epsilon_-(1-\epsilon_+), \\
&P(\hat{z}_{ij} = y_{ij}\ \text{[True outcome (TO)]}) &&= (1-\epsilon_+)(1-\epsilon_-), \\
&P(\hat{z}_{ij} = -y_{ij}\ \text{[Inverse outcome (IO)]}) &&= \epsilon_+\epsilon_-. \\
\end{aligned}
\right.
\end{equation}
Correspondingly, the probabilities of obtaining the correct Copeland score $P(R_{\bullet},\text{TO})$, are now
{\small
\begin{equation}
\begin{aligned}
P(R_1,\text{TO}) &= (1-\epsilon_+)^{N-1}(1-\epsilon_-)^{N-1}, \\
P(R_2,\text{TO}) &= (1-\epsilon_+)^{N-1}(1-\epsilon_-)^{N-1} + (1-\epsilon_+)^{N-3}(1-\epsilon_-)^{N-3}C_1^1C_{N-2}^1(\epsilon_+\epsilon_-)^2, \\
P(R_3,\text{TO}) &= (1-\epsilon_+)^{N-1}(1-\epsilon_-)^{N-1} + (1-\epsilon_+)^{N-3}(1-\epsilon_-)^{N-3}C_2^1C_{N-3}^1(\epsilon_+\epsilon_-)^2 + (1-\epsilon_+)^{N-5}(1-\epsilon_-)^{N-5}C_2^2C_{N-3}^2(\epsilon_+\epsilon_-)^4, \\
...\\
P(R_m,\text{TO}) &= (1-\epsilon_+)^{N-1}(1-\epsilon_-)^{N-1} + (1-\epsilon_+)^{N-3}(1-\epsilon_-)^{N-3}C_{m-1}^1C_{N-m}^1(\epsilon_+\epsilon_-)^2 + \\
&(1-\epsilon_+)^{N-5}(1-\epsilon_-)^{N-5}C_{m-1}^2C_{N-m}^2(\epsilon_+\epsilon_-)^4 + ... + (1-\epsilon_+)^{N-1-2(m-1)}(1-\epsilon_-)^{N-1-2(m-1)}C_{m-1}^{m-1}C_{N-m}^{m-1}(\epsilon_+\epsilon_-)^{2(m-1)},\\
...\\
P(R_{N-1},\text{TO}) &= (1-\epsilon_+)^{N-1}(1-\epsilon_-)^{N-1} + (1-\epsilon_+)^{N-3}(1-\epsilon_-)^{N-3}C_1^1C_{N-2}^1(\epsilon_+\epsilon_-)^2, \\
P(R_N,\text{TO}) &= (1-\epsilon_+)^{N-1}(1-\epsilon_-)^{N-1},\\
\end{aligned}
\end{equation}
}

with the second half $m > (N+1)/2$ mirroring the first half $m \leq (N+1)/2$ via the substitution $m-1 \rightarrow N-m$. 

This formulation nonetheless does not separate $\epsilon_+$ and $\epsilon_-$, which always appear together in $(1-\epsilon_+)(1-\epsilon_-)$ or $\epsilon_+\epsilon_-$. To separate the two error terms for possible estimation, we can conduct repeated comparisons between each pair of objects. 

Consider repeated comparisons between $i$ and $j$ in a series of instances denoted by $K$ (e.g., $K = [+,-,+,-,...]$) in which there are $k_+$ instances of $+$, i.e., object $i$ is in the first place and the outcome is $\hat{y}_{ij,t}$ (the $t$-th such case), and $k_-$ instances of $-$, i.e., object $j$ is in the first place and the outcome is $\hat{y}_{ji,t}$ (the $t$-th such case). Consider separate error terms $\epsilon_+$, $\epsilon_-$. Define the observation matrix $\mathbf{W}=\{\hat{w}_{ij}\}$ (antisymmetric, $\hat{w}_{ij} = -\hat{w}_{ji}$), where
\begin{equation}
    \hat{w}_{ij} = \hat{w}_{ij}(K) \approx \hat{w}_{ij}(k_+,k_-) = \frac{\sum^{k_+} \hat{y}_{ij,\bullet} - \sum^{k_-} \hat{y}_{ji,\bullet}}{k_+ + k_-}.
\end{equation}
This assumes that the sequence of $K$ breaks down to the two counts $k_+$ and $k_-$. The two sums in the numerator are taken over the two categories of outcomes, respectively. $\mathbf{W}=\{\hat{w}_{ij}\}$ reduces to $\mathbf{Z}=\{\hat{z}_{ij}\}$ when $k_+=k_- = 1$.

Unlike at the uniform error, where there are only three possible values (1, 0, -1) for $\hat{z}_{ij}$, now there are $1 + k_+ + k_-$ possible values for $\hat{w}_{ij}$: $1, \frac{k_+ + k_- - 2}{k_+ + k_-}, \frac{k_+ + k_- - 4}{k_+ + k_-}, \frac{k_+ + k_- - 6}{k_+ + k_-}, ..., -1$. The probabilities for observing these values at $\hat{w}_{ij}$ are given by (suppose the ground truth $y_{ij} = 1$)
\begin{equation}
P(\hat{w}_{ij} = \frac{k_+ + k_- - 2m}{k_+ + k_-}) = \sum^{m_+ + m_- = m}_{m_+ \in [0,k_+]; m_- \in [0,k_-]} C_{k_+}^{m_+}C_{k_-}^{m_-} (1-\epsilon_+)^{k_+ - m_+}(1-\epsilon_-)^{k_- - m_-}\epsilon_+^{m_+}\epsilon_-^{m_-};
\end{equation}
in particular, 
\begin{equation}
\left\{
\begin{aligned}
&P(\hat{w}_{ij} = y_{ij}\ \text{[True outcome (TO)]}) &&= (1-\epsilon_+)^{k_+}(1-\epsilon_-)^{k_-}, \\
&P(\hat{w}_{ij} = -y_{ij}\ \text{[Inverse outcome (IO)]}) &&= \epsilon_+^{k_+}\epsilon_-^{k_-}. \\
\end{aligned}
\right.
\end{equation}

We calculate Copeland scores from the matrix $\mathbf{W}$:
\begin{equation}
S^{W}_{Copeland}(i) = \sum_j^{N-1} \hat{w}_{ij},
\end{equation}
and compute the probabilities of having the correct Copeland score $P(R_{\bullet},\text{TO})$:
{\small
\begin{equation}
\begin{aligned}
P(R_1,\text{TO}) &= (1-\epsilon_+)^{k_+(N-1)}(1-\epsilon_-)^{k_-(N-1)}, \\
P(R_2,\text{TO}) &= (1-\epsilon_+)^{k_+(N-1)}(1-\epsilon_-)^{k_-(N-1)} + (1-\epsilon_+)^{k_+(N-3)}(1-\epsilon_-)^{k_-(N-3)}C_1^1C_{N-2}^1(\epsilon_+^{k_+}\epsilon_-^{k_-})^2, \\
P(R_3,\text{TO}) &= (1-\epsilon_+)^{k_+(N-1)}(1-\epsilon_-)^{k_-(N-1)} + (1-\epsilon_+)^{k_+(N-3)}(1-\epsilon_-)^{k_-(N-3)}C_2^1C_{N-3}^1(\epsilon_+^{k_+}\epsilon_-^{k_-})^2 + \\
&(1-\epsilon_+)^{k_+(N-5)}(1-\epsilon_-)^{k_-(N-5)}C_2^2C_{N-3}^2(\epsilon_+^{k_+}\epsilon_-^{k_-})^4, \\
...\\
P(R_m,\text{TO}) &= (1-\epsilon_+)^{k_+(N-1)}(1-\epsilon_-)^{k_-(N-1)} + (1-\epsilon_+)^{k_+(N-3)}(1-\epsilon_-)^{k_-(N-3)}C_{m-1}^1C_{N-m}^1(\epsilon_+^{k_+}\epsilon_-^{k_-})^2 + \\
&(1-\epsilon_+)^{k_+(N-5}(1-\epsilon_-)^{k_-(N-5}C_{m-1}^2C_{N-m}^2(\epsilon_+^{k_+}\epsilon_-^{k_-})^4 + ... + \\
&(1-\epsilon_+)^{k_+(N-1-2(m-1))}(1-\epsilon_-)^{k_-(N-1-2(m-1))}C_{m-1}^{m-1}C_{N-m}^{m-1}(\epsilon_+^{k_+}\epsilon_-^{k_-})^{2(m-1)},\\
...\\
P(R_N,\text{TO}) &= (1-\epsilon_+)^{k_+(N-1)}(1-\epsilon_-)^{k_-(N-1)}.\\
\end{aligned}
\end{equation}
}

This formulation separates $\epsilon_+$ and $\epsilon_-$ whenever $k_+\neq k_-$, and the two rates can be estimated accordingly. The behavior of $dP(R_m,\text{TO})/dN$ here at $\mathbf{W}$ is similar to that of at $\mathbf{Z}$; both $\mathbf{W}$ and $\mathbf{Z}$ are antisymmetric. 

\subsection*{Estimation of error}

The $S^{Z/W}_{Copeland}$ scores computed from the observed $\mathbf{Z}$ or $\mathbf{W}$ matrix do not follow the perfect sequence $N-1$, $N-3$, $N-5$, ..., $5-N$, $3-N$, $1-N$ due to the error in the output. We calculate its deviation from the perfect sequence, defined as the sum of the distances between elements in the two (ordered) sequences, denoted as $\Delta_S$. $\Delta_S$ reveals the loss of scalability at pairwise comparison: $\Delta_S$ grows as $N$ increases, at a speed inferior to the case of an all-zero $\mathbf{Z}$ matrix, in which case there is the closed form $\Delta_S = N(N+1)/2$. 

The $\Delta_S-N$ plot helps yield the estimates for $\epsilon/\epsilon_{+,-}$. For uniform error, at a specific value of $\epsilon$, (\ref{eq-p}) realizes a set of probabilities for elements in $\mathbf{Z}$ to take values of -1, 0, 1, respectively; the resulting $\mathbf{Z}$ matrix helps compute the $\Delta_S$ at this $\epsilon$ level. The error-free case ($\epsilon=0$) reproduces the perfect sequence ($\Delta_S = 0$); as $\epsilon$ increases, elements in $\mathbf{Z}$ get more random, and the deviation $\Delta_S$ increases. Computing for different $N$ and fitting the $\Delta_S-N$ plot from the measured $\mathbf{Z}$ to plots from $\epsilon$-specific $\mathbf{Z}$s reveals the best-fit error level. 

For positional bias, similarly, a set of $\epsilon_+,\epsilon_-$ results in a simulated $\Delta_S-N$ curve calculated from $\mathbf{W}$. The ground-truth $\mathbf{W}$ is computed from the LLM output, and the best-fit $\epsilon_+,\epsilon_-$ can be estimated accordingly. In practice, a subset of observations from the repeated comparisons can be used to cross-validate the error estimates: consider $k_+^s \le k_+$ and $k_-^s \le k_-$, and use the corresponding $\mathbf{W}^{k_+^s, k_-^s}$ to obtain the estimates $\epsilon_+^{k_+^s, k_-^s},\epsilon_-^{k_+^s, k_-^s}$; check the consistency of estimates across the values of $k_+^s, k_-^s$ and report the average results.

\subsection*{The Bradley-Terry model}

The Bradley-Terry (BT) model \citep{BT1952} is widely used for efficiently constructing a ranking over objects from their pairwise comparison \citep[e.g.,][]{HWet2006,MM2008,TF2012}. It is adopted by \citep{LRet2024} to turn the LLM preferences into an effective ranking using a subset of the comparison outcomes. The model assumes relative scores for the compared objects and constructs a maximum likelihood estimator to yield the optimal scores that meet the comparison outcomes, which can be solved with Zermelo's algorithm \citep{Z1929} and notably its recent faster variant \citep{N2023}.

Assume a set of scores for the $N$ objects, $s_1, s_2, ..., s_N$. For the original (unbiased) BT model, the probability of $i$ better than $j$ depends on the scores of $i$ and $j$ via the sigmoid function $\sigma(x) = 1/(1+e^{-x})$: $P(i \text{ better than }j) = \sigma(s_i-s_j)=e^{s_i}/(e^{s_i}+e^{s_j})=\pi_i/(\pi_i+\pi_j)$, writing $e^{s_i}$ as $\pi_i$ (following \citep{N2023}). The probability of observing $\hat{y}_{ij}\in[0,1]$ is then $P(\hat{y}_{ij}) = \sigma(s_i-s_j)^{\hat{y}_{ij}}[1-\sigma(s_i-s_j)]^{1-\hat{y}_{ij}} = [\pi_i/(\pi_i+\pi_j)]^{\hat{y}_{ij}}[\pi_j/(\pi_i+\pi_j)]^{1-\hat{y}_{ij}}$. Best-fit scores $\hat{s}_{1:N}$ are estimated from maximizing the complete likelihood of observing all the outcomes $\hat{y}_{ij}$, i.e., $\hat{s}_{1:N} = argmax_{s_{1:N}} \Pi_{i,j}P(\hat{y}_{ij})$. 

Consider the error in the LLM output and incorporate bias into the BT model. The idea is adding a bias shift, following \citep{LRet2024}: $(s_i-s_j) \rightarrow (s_i-s_j-\epsilon)$. In this case, 
\begin{equation}
    P_{\dagger}(i \text{ better than }j,\ \epsilon) = \sigma(s_i-s_j-\epsilon) = \frac{e^{s_i}}{e^{s_i}+e^{\epsilon}e^{s_j}} = \frac{\pi_i}{\pi_i+e^{\epsilon}\pi_j}.
\end{equation}
Best-fit $\hat{s}_{1:N}$ are estimated with
\begin{equation}
    \hat{s}_{1:N} = argmax_{s_{1:N}} \Pi_{i,j}P_{\dagger}(\hat{y}_{ij},\epsilon).
\end{equation}
See Supplementary Materials E for the iteration procedure of the biased BT model. An alternative idea is adding a multiplicative term to the sigmoid, $P(x) = \sigma[(1-\epsilon)x]$; yet this formulation is not effective as we consider the ranking of the objects instead of their absolute scores. The BT model applies to repeated comparisons between object pairs; the strength matrix \citep{N2023} $\mathbf{X}=\{\hat{x}_{ij}\}$ where $\hat{x}_{ij}$ is the count of $i$ beating $j$.

We can use this biased BT model to estimate the error rate $\epsilon$ by adding constraints to the estimated scores. For example, we can assume the obtained $\hat{s}_{1:N}$ to have a small/large spread and then search for the optimal $\epsilon$ that minimizes/maximizes the spread of $\hat{s}_{1:N}$. Moreover, with a given $\epsilon$, e.g., estimated from our model, we can obtain the scores $\hat{s}_{1:N}$ and the corresponding text ranking. 

\section*{Materials}

\textbf{Text materials.} Five types of text input are used in the experiment (Table 1; see sample texts in Supplementary Materials A): (i) pseudo-word paragraphs, (ii) pseudo paragraphs, (iii) advertising slogans, (iv) short poems, (v) academic abstracts. For (i), we generate 100 paragraphs of pseudo-word (void of meaning) sentences. For (ii), we generate 100 paragraphs with random English words. For (iii), we consider 100 popular commercial slogans (with one-line introductions) \footnote{https://advertisingnews.com/popular-slogans/}. For (iv), we consider 100 famous short poems on PoetrySoup\footnote{https://www.poetrysoup.com/famous/poems/short/top\_100\_famous\_short\_poems.aspx}. For (v), we consider 100 abstracts of articles (across disciplines) published on Science Advances in May 2025. The five types of text have different average numbers of words and average numbers of tokens (converted from the text string) when fed into the LLM.

\begin{table}[h!]
\centering
\resizebox{0.7\textwidth}{!}
{
\begin{tabular}{cccc}

\multicolumn{4}{l}{\textit{text materials}} \\ 
\hline
\textbf{Type} & $N$ & \textbf{Ave. No. Words} & \textbf{Ave. No. Tokens} \\
\hline
pseudo-word paragraphs & 100 & 100 & 278 \\
pseudo paragraphs & 100 & 100 & 133 \\
advertising slogans & 100 & 17 & 23 \\
short poems & 100 & 49 & 66 \\
academic abstracts & 100 & 157 & 219 \\
\hline
\end{tabular}
}
\caption{\textbf{Text materials used in the experiment.} No.: number. Ave.: average.}
\label{tab}
\end{table}

\textbf{LLMs.} We adopt six popular LLMs: ChatGPT (4.1 mini), Claude (3.5 sonnet), Deepseek (V3-30324), Gemini (2.5 Flash), Grok (3 mini), and Qwen (2.5-72B). The stochastic level (temperature) is set uniformly at 0.1. The following zero-short prompt is used: \textit{(system) You are a senior [text type] evaluation expert with rich experience in literary appreciation and judgment. Please conduct comprehensive text evaluations based on quality, creativity, expression effectiveness, and other dimensions. Only respond with 1 or 2; no explanation needed. (user) Compare the following two [text type] and indicate which one is better. Output only the number: 1 if Text 1 is better, 2 if Text 2 is better. Text 1: $\{$text$\}$. Text 2: $\{$text$\}$.} Three prompt variants are considered in the robustness checks (Supplementary Materials B).

\section*{Results}

\begin{figure}[h]
  \centering
  \vspace{-0.4cm}
  \includegraphics[width=6.5in]{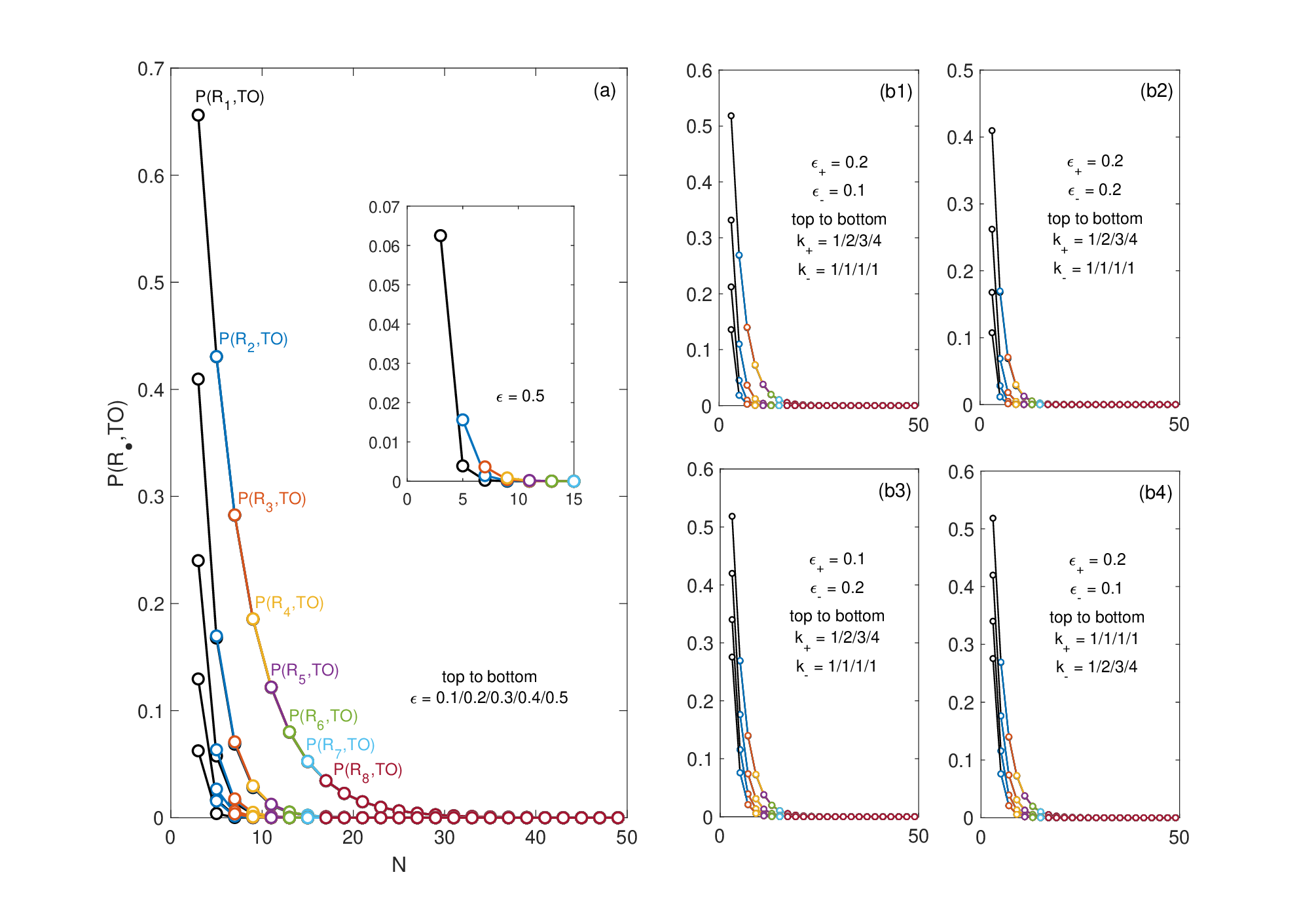}
  \vspace{-0.7cm}
  \caption{\textbf{Scalability of the Copeland ranking based on pairwise comparisons.} Showing the probability of the $\#m$-ranking text obtaining the correct Copeland score, $P(R_{m},TO)$. $m = 1$ to 8. (a) Uniform error. $S_{Copeland}^{Z}$ calculated on $\mathbf{Z}$. $\epsilon = 0.1$ to 0.5. Inset: $\epsilon = 0.5$. (b1)-(b4) Positional bias. $S^{W}_{Copeland}$ calculated on $\mathbf{W}$. Values of $\epsilon_+, \epsilon_-$ and $k_+, k_-$ are shown on the plots.}
  \label{fig1}
\end{figure}

\textbf{Scalability of text ranking based on pairwise comparisons.} We visualize the probability of the $\#m$-ranking text obtaining the correct Copeland score, $P(R_{m},TO)$, constructed from LLMs' pairwise preferences. Show the results for $m = 1$ to 8. For uniform error, the scores $S_{Copeland}^{Z}$ are calculated on matrix $\mathbf{Z}$; simulations verify that $dP(R_m,\text{TO})/dN < 0$ for different $\epsilon$ and $m$ (Figure \ref{fig1}(a); $\epsilon = 0.1$ to 0.5). $P(R_{\bullet},TO)$ approaches $P(R_{1},TO)$ when $\epsilon$ is small, as the terms with the multipliers of $\epsilon$ vanish (see equation (8)); the $P(R_{2-8},TO)-N$ curves thus overlap the $P(R_{1},TO)-N$ curve. The difference between $P(R_{1},TO)$, $P(R_{2},TO)$, and $P(R_{3},TO)$ can be seen, for example, at a large $\epsilon=0.5$ (Figure \ref{fig1}(a) inset). 

For repeated comparisons, $S^{W}_{Copeland}$ are calculated on $\mathbf{W}$, and $P(R_{\bullet},TO)$ are dependent on the counts of comparisons $k_+$, $k_-$ as well as the error rates $\epsilon_+$, $\epsilon_-$. Similar to uniform error, $dP(R_m,\text{TO})/dN < 0$ for different $m$ and combinations of $k_+$, $k_-$, $\epsilon_+$, $\epsilon_-$ (Figure \ref{fig1}(b1)-(b4); four sets of parameter values are shown). Consistently, switching the values of $k_{+/-}$ and $\epsilon_{+/-}$ simultaneously retains the outcome (Figure \ref{fig1}(b3)-(b4)) due to the interchangeable role of these two sets of parameters in indicating the positional bias. Overall, simulation results confirm that for sufficiently large $N$, $dP(R_m,\text{TO})/dN < 0$ for any $m$ and error rates $\epsilon/\epsilon_{+,-}$: the Copeland ranking constructed from the observation matrix $\mathbf{Z}$ or $\mathbf{W}$ are NOT scalable, i.e., the ranking gets more error-prone when more objects are compared.

\begin{figure}[h]
  \centering
  \vspace{-0.4cm}
  \includegraphics[width=6.5in]{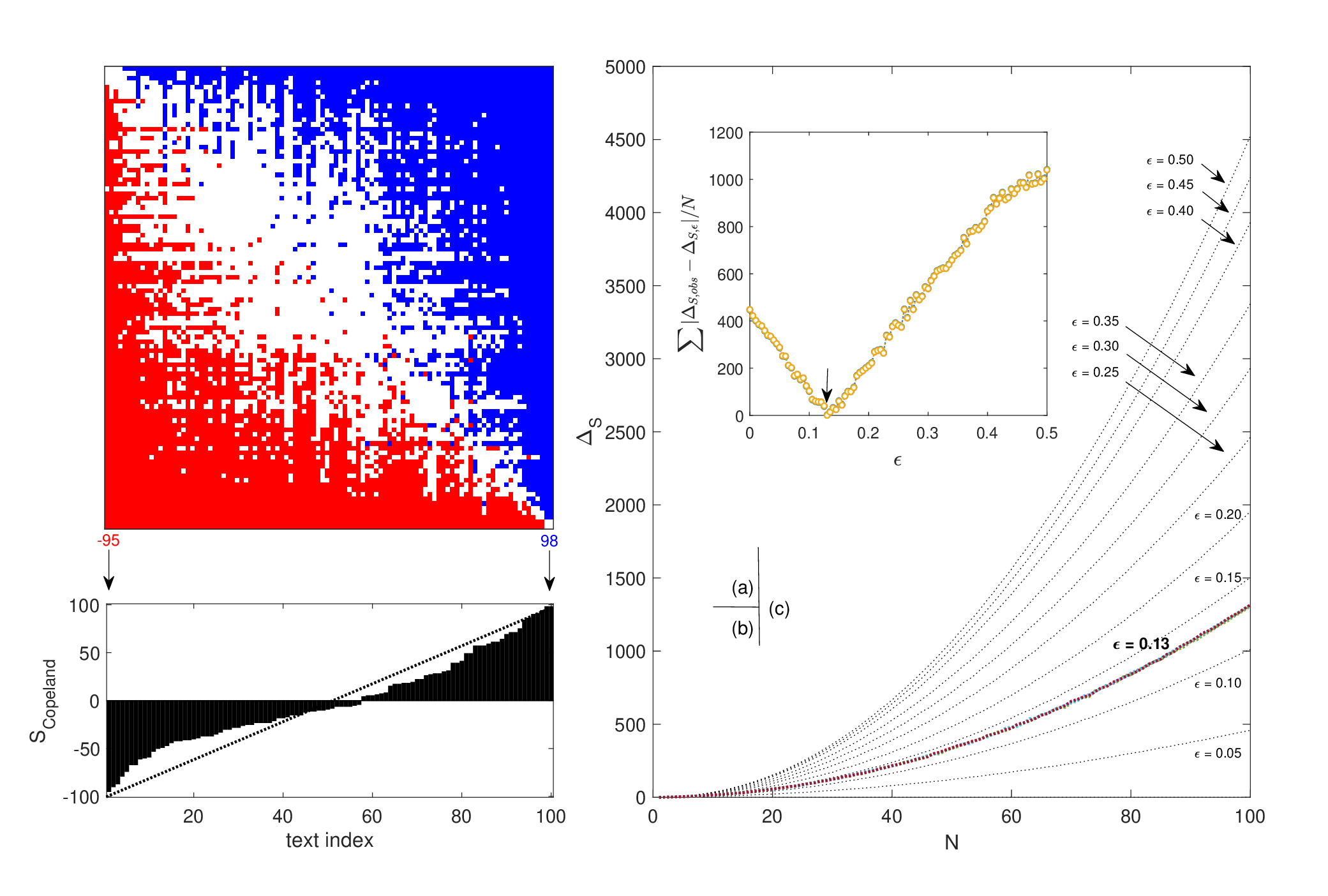}
  \vspace{-0.9cm}
  \caption{\textbf{Estimation of LLMs' error rate from the output preferences (uniform error $\epsilon$).} Input text: short poems. LLM: ChatGPT. (a) Observation matrix $\mathbf{Z}$ constructed from pairwise preferences $\hat{y}_{ij}$. Red/Blue/White: -1/0/1 entries. (b) Copeland scores $S_{Copeland}^Z$ calculated on $\mathbf{Z}$. Highest/lowest scores (98/-95) are obtained at the corresponding columns of $\mathbf{Z}$. Dash line: perfect scores. The area between the dash line and the observed $S_{Copeland}^Z$ is the difference $\Delta_{S,obs}$. (c) Fitting the $\Delta_{S,obs}-N$ curve. Light curves: synthetic $\Delta_{S, \epsilon}-N$ calculated with $\epsilon\in[0.05,0.5]$ at interval 0.05. Thick curves: the empirical $\Delta_{S,obs}-N$. Inset: grid search ($\epsilon\in[0,0.5]$ at interval 0.005) minimizing the summed distance between the empirical and the synthetic curves. Best-fit $\epsilon = 0.130$, i.e., a uniform error rate of 13.0$\%$. Results from three empirical runs of comparison largely overlap (different colors at the thick curves and in the inset), which have the commutativity scores $S_{com} = 39.01\%, 38.95\%, 39.03\%$, respectively; three instances of estimate are identical.}
  \label{fig2}
\end{figure}

\textbf{Estimation of LLMs' error at text comparison.} We demonstrate the estimation with short poems as the input text and ChatGPT as the LLM, first at uniform error $\epsilon$, i.e., conducting two comparisons between each text pair with either text placed in the front and combining the two preferences. The observation matrix $\mathbf{Z}$ is constructed from the preferences $\hat{y}_{ij}$ and the Copeland scores $S_{Copeland}^Z$ are calculated on $\mathbf{Z}$ (Figure \ref{fig2}a, b). We sample $n$ texts from the ensemble, calculate the $S_{Copeland}^Z$ on the corresponding sub-matrix of $\mathbf{Z}$, and sum the difference between the obtained and the perfect Copeland score sequence, $\Delta_S$. For each $2\le n\le N$, we conduct 200 runs of the sampling and compute the mean $\Delta_S$. We fit the empirical $\Delta_{S,obs}-N$ curve to the $\Delta_{S,\epsilon}-N$ curves (Figure \ref{fig2}c) calculated with synthetic $\mathbf{Z}$ generated at different levels of $\epsilon\in[0,0.5]$, using grid search at interval 0.005. Each $\Delta_{S,\epsilon}$ averages over 10 synthetic $\mathbf{Z}$ at a certain $\epsilon$.

We collect the outcomes of three runs of comparison and conduct the estimation three times; one instance of the matrix $\mathbf{Z}$ and the calculated Copeland scores $S_{Copeland}^Z$ is illustrated (Figure \ref{fig2}a, b). Fitting the empirical $\Delta_{S,obs}-N$ to synthetic $\Delta_{S}-N$ curves (Figure \ref{fig2}c) reveals a clear best-fit $\epsilon = 0.130$ with near-zero misfit (Figure \ref{fig2}c inset), i.e., a uniform error rate of 13.0$\%$. The three instances of the outcome are almost identical, having the commutativity scores $S_{com} = 39.01\%, 38.95\%, 39.03\%$ and the same best-fit $\epsilon$. 

We then demonstrate the estimation at positional bias with repeated comparisons. We consider a comparison order sequence $K = [+,-,+,-,+,-]$, reduced to $k_{+/-} = 3$. We show the results at $\mathbf{W}^{1,1}$ to $\mathbf{W}^{3,3}$; $\mathbf{W}^{1,1}$ reduces to the $\mathbf{Z}$ at uniform error. We look for the best-fit $\epsilon_+,\epsilon_-\in[0,0.5]$ using 2-D grid search, minimizing the summed distance between the empirical $\Delta_{S,obs}-N$ and the synthetic $\Delta_{S}-N$ curves, as before.

\begin{figure}[h]
  \centering
  \vspace{-0.3cm}
  \includegraphics[width=5in]{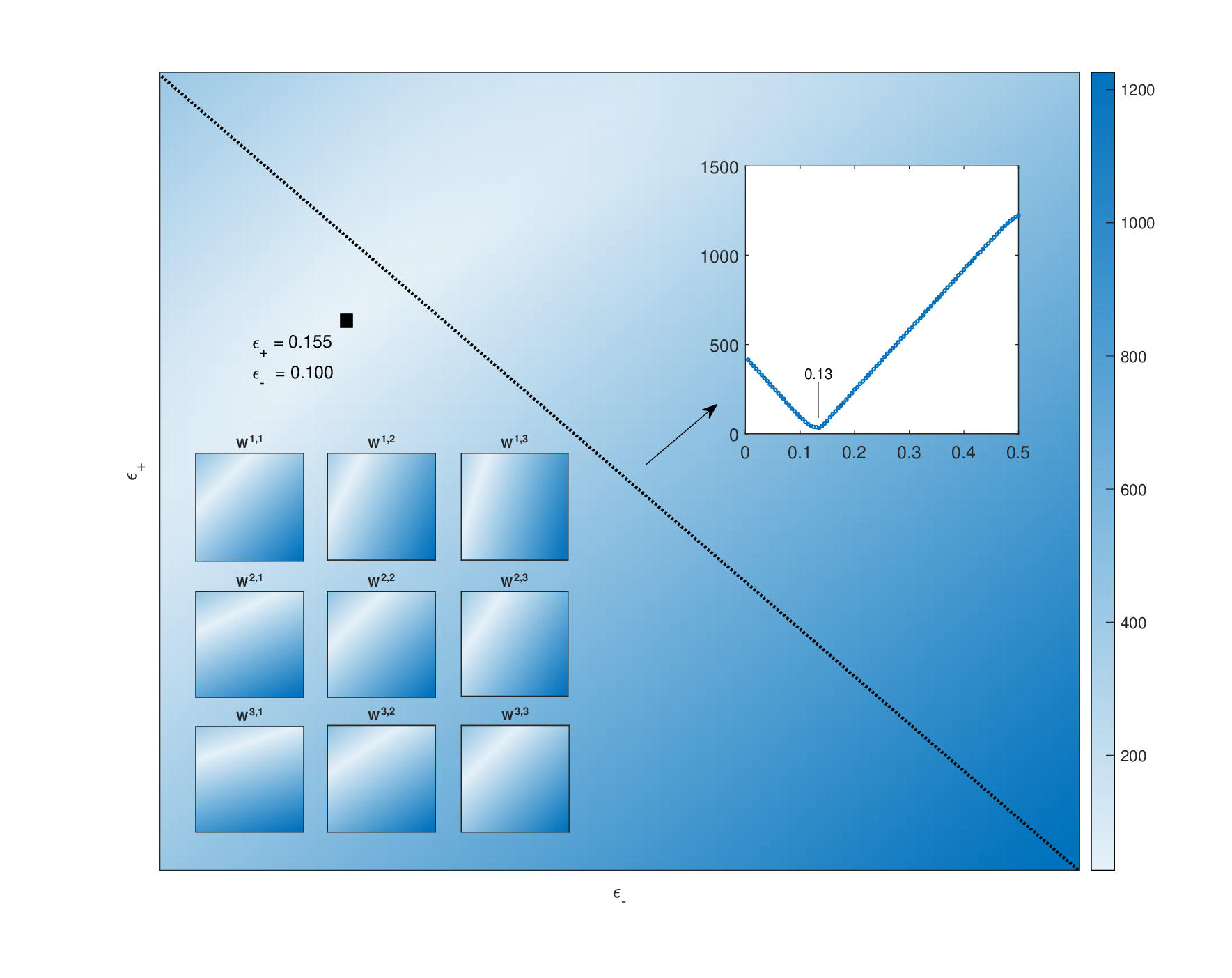}
  \vspace{-0.7cm}
  \caption{\textbf{Estimation of LLMs' error rate from the output preferences (positional bias $\epsilon_{+/-}$).} Input text: short poems. LLM: ChatGPT. Minimizing the summed distance between the empirical $\Delta_{S,obs}-N$ and the synthetic $\Delta_{S}-N$ curves. 2-D grid search $\epsilon_+,\epsilon_-\in[0,0.5]$ at interval 0.005. The main heatmap shows the result averaging the nine cases of $\mathbf{W}^{1,1}$ to $\mathbf{W}^{3,3}$ (left inset). Diagonal elements (right inset) concur with the curve at uniform error. Best-fit $\epsilon_+ = 0.155, \epsilon_- = 0.100$ on the average $\mathbf{W}$.}
  \label{fig3}
\end{figure}

Figure \ref{fig3} shows the estimates of $\epsilon_+,\epsilon_-$ at the nine cases $\mathbf{W}^{1,1}$ to $\mathbf{W}^{3,3}$ (left inset) and the average result (main heatmap), whose diagonal elements, corresponding to $\epsilon_+ =\epsilon_- = 0:0.005:0.5$ (right inset), concur with the curve at uniform error (Figure \ref{fig2} inset). Best-fit $\epsilon_+ = 0.155,\epsilon_- = 0.100$ on the average $\mathbf{W}$. The best-fits slightly vary under different values of $k_+,k_-$. One possibility is that instances of the LLM's output from a series of comparisons are not \textit{independent}, which is assumed when we calculate the observing probabilities ((3) and (18)): for the error of the output, there might be another positional factor regarding its place ($t$) in the comparison series ($K$), i.e., $\epsilon = \epsilon(K,t)$. This could be studied with $K$-specific $\mathbf{W}^{K}$ in future works.

We conduct the above error estimation with five types of text input (pseudo-word paragraphs, pseudo paragraphs, advertising slogans, short poems, academic abstracts) at six LLMs (ChatGPT, Claude, DeepSeek, Gemini, Grok, Qwen), under matrix $\mathbf{Z}$ (uniform error) or $\mathbf{W}$ (positional bias, averaging over $k_+,k_-\in[1,3]$). 

Results are shown in Figure \ref{fig4} (see Table S1 in the Supplementary Materials for values). Across the six LLMs, at pseudo-word paragraphs and pseudo paragraphs, the estimated $\epsilon, \epsilon_{+/-}$ fall in [0.2, 0.5], while at slogans, poems, and abstracts, the estimates fall in [0, 0.2] (except for at slogans and abstracts under Gemini). This is consistent with the idea that LLMs are more error-prone when used to compare meaningless content, where a clear winner cannot be identified, than when used to compare meaningful content. At pseudo-word paragraphs and pseudo paragraphs, Qwen, Gemini, and Claude have the average error closer to 0.5, which is desirable, than Grok, DeepSeek, and ChatGPT. At the other text types, the average error is smaller at Qwen, Grok, and Claude than at ChatGPT, DeepSeek, and Gemini. Overall, in this experiment, Qwen obtained the best performance among the six LLMs, followed by Claude, Grok, then ChatGPT and DeepSeek; Gemini exhibited the worst performance. In most cases, the estimates of $\epsilon_{+/-}$ are not far from the estimate of $\epsilon$, especially at the three meaningful text types. There, the deviation of $\epsilon_{+/-}$ from uniform $\epsilon$ is the largest under Gemini, followed by ChatGPT; the deviation is trivial at the other four LLMs.

\begin{figure}[h]
  \centering
  \includegraphics[width=6.5in]{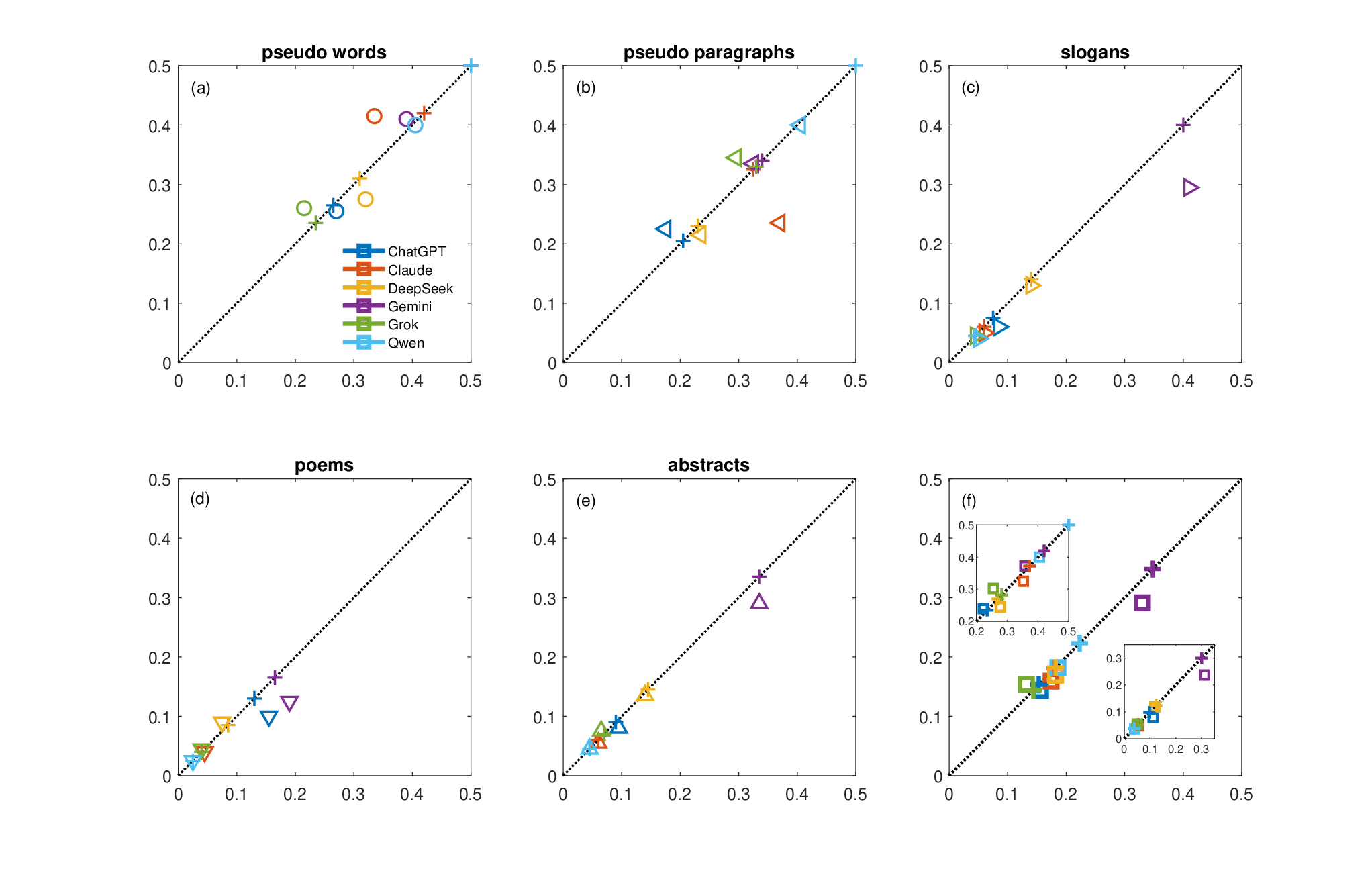}
  \vspace{-0.5cm}
  \caption{\textbf{Main estimation results.} Grid search for $\epsilon, \epsilon_{+/-}\in[0,0.5]$ at interval 0.005. Crosses: $\epsilon$ estimates at uniform error (using matrix $\mathbf{Z}$). Other markers: $\epsilon_{+/-}$ estimates at positional bias (using matrix $\mathbf{W}$). (a)-(e): Five types of text input. (f) Average estimates: over the five types (main), over pseudo-word and pseudo paragraphs (upper inset), over the other three types (lower inset).}
  \label{fig4}
  \vspace{-0.3cm}
\end{figure}

We test three ideas to vary the baseline prompt (Supplementary Materials B): (V1) the text type is not specified; (V2) the prompt is presented in another language (Chinese); (V3) the system prompt is exempted. The variations of $S_{com}$ and $\epsilon$ are summarized in Table 2 (see Table S2 in the Supplementary Materials for values). Overall, under these prompt variants where the original prompt is uniformly weakened in one way or another, $S_{com}$ and the estimated $\epsilon$ increase, i.e., greater error rates are suggested. This is consistent with the finding that LLMs' performance differs across languages \citep{LNet2023} and levels of specificity \citep{MGet2023}. Of the six LLMs, Claude showed the highest robustness to the variation of prompts, followed by Gemini; interestingly, the error decreased at these two LLMs under certain prompt variants. Qwen demonstrated the least robustness, despite its good performance in the main result (Figure \ref{fig4}). In general, it is fair to note that the error estimates in this experiment are not sufficiently robust to the variation of prompts. Considering both the error rates and the robustness to prompts, Claude outperformed the other LLMs.

\begin{table}[htbp]
\centering
\resizebox{0.85\textwidth}{!}
{
\begin{tabular}{c|c|c|c|c|c|c|c|c}
\hline
& \multicolumn{4}{c|}{$\Delta S_{com}$} & \multicolumn{4}{c}{$\Delta \epsilon$} \\ \hline 
\textbf{LLM} & \textbf{V1} & \textbf{V2} & \textbf{V3} & \textbf{mean} & \textbf{V1} & \textbf{V2} & \textbf{V3} & \textbf{mean}  \\ \hline 
ChatGPT & +18.6\% & +24.0\% & +37.6\% & +26.7\% & +27.6\% & +42.8\% & +66.8\% & +45.7\% \\
Claude & +5.4\% & +9.9\% & -5.9\% & +3.1\% & +6.2\% & +10.4\% & -9.0\% & +2.5\% \\
DeepSeek & +31.2\% & +2.1\% & +58.5\% & +30.6\% & +63.4\% & +5.3\% & +94.9\% & +54.5\% \\
Gemini & -30.7\% & -21.9\% & +13.2\% & -13.1\% & -35.8\% & -29.1\% & +12.1\% & -17.6\%\\
Grok & +48.8\% & +22.1\% & +21.7\% & +30.9\% & +42.7\% & +18.3\%& +21.1\% & +27.4\% \\
Qwen & +44.7\% & +68.4\% & +8.0\% & +40.4\% & +57.6\% & +113.1\% & +8.2\% & +66.3\%  \\
\hline
average $|\Delta|$ & 29.9\% & 24.7\% & 24.2\% & - & 38.9\% & 36.5\% & 35.4\% & - \\
\hline
\end{tabular}
}
\caption{\textbf{Variation of $S_{com}$ and $\epsilon$ under prompt variants.} Each entry averages over the five text types.}
\end{table}

Our estimates of $\epsilon$ align with the commutativity scores (Figure \ref{fig6}a): when $S_{com}$ is small/large, the estimated $\epsilon$ is small/large, albeit the lack of strict monotonicity. Interestingly, the ratio of $S_{com}/\epsilon$ increases as $\epsilon$ decreases: at pseudo-word paragraphs and pseudo paragraphs (as well as two other cases under Gemini) where $\epsilon>0.2$, $S_{com}/\epsilon<2.5$, while $S_{com}/\epsilon>2.9$ when $\epsilon<0.2$. This suggests that our estimates reflect the error of LLMs better than the commutativity scores.

\begin{figure}[h]
  \centering
  \includegraphics[width=6.5in]{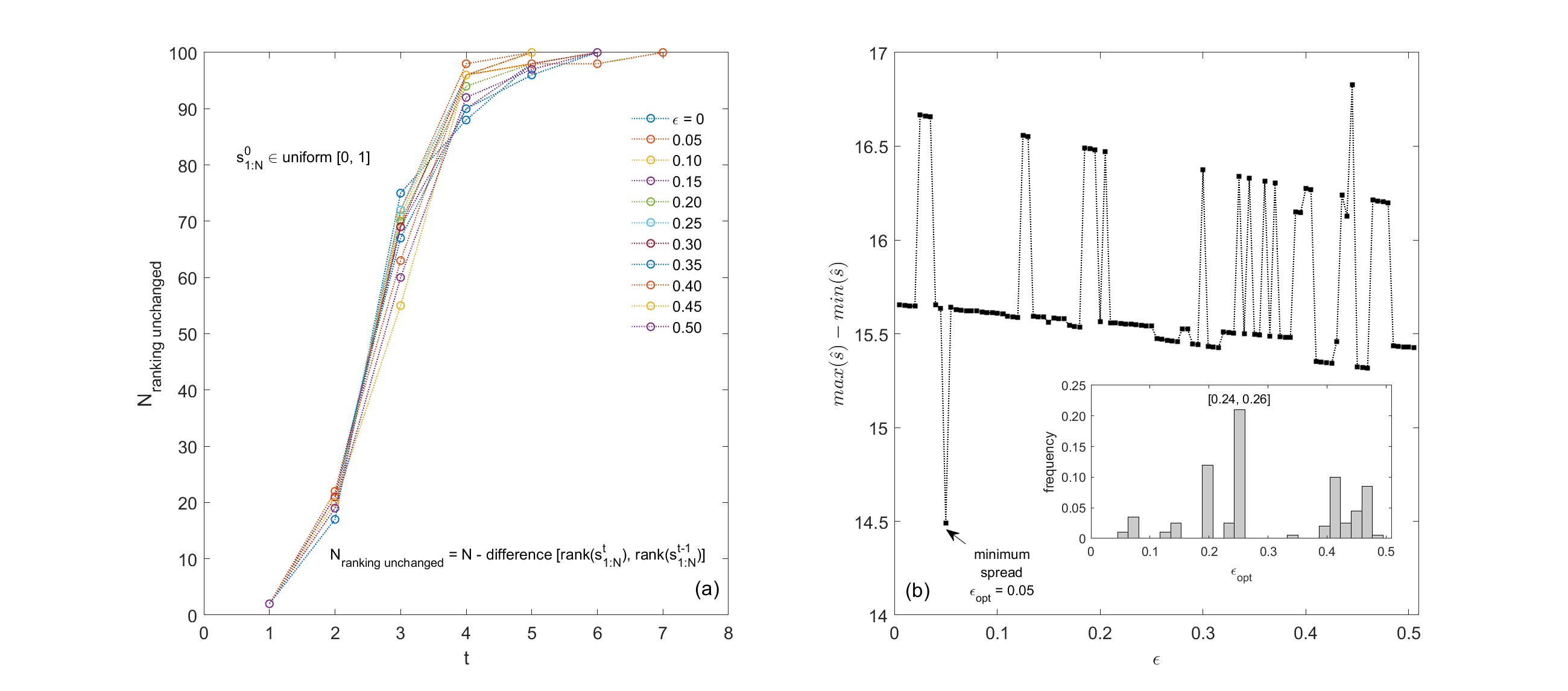}
  \caption{\textbf{Estimation with the biased Bradley-Terry model.} Estimation of $\epsilon$ with $\mathbf{X}^{3,3}$. Input: short poems. LLM: ChatGPT. (a) Convergence of iterations. Initial scores $\hat{s}_{1:N}\in uniform\ [0,1]$. Y-axis: number of texts with unchanged ranking across one iteration. X-axis: number of iterations. (b) Grid search ($\epsilon$ at interval 0.005) minimizing the spread of estimated scores $\hat{s}$. Minimum spread obtained at $\epsilon_{opt}=0.05$. Inset: frequency of $\epsilon_{opt}$ across 200 random seeds for the initial scores (resolution 0.02, i.e., 25 bins within [0, 0.5]).}
  \label{fig5}
\end{figure}

\begin{figure}[htbp!]
  \centering
  \includegraphics[width=6.5in]{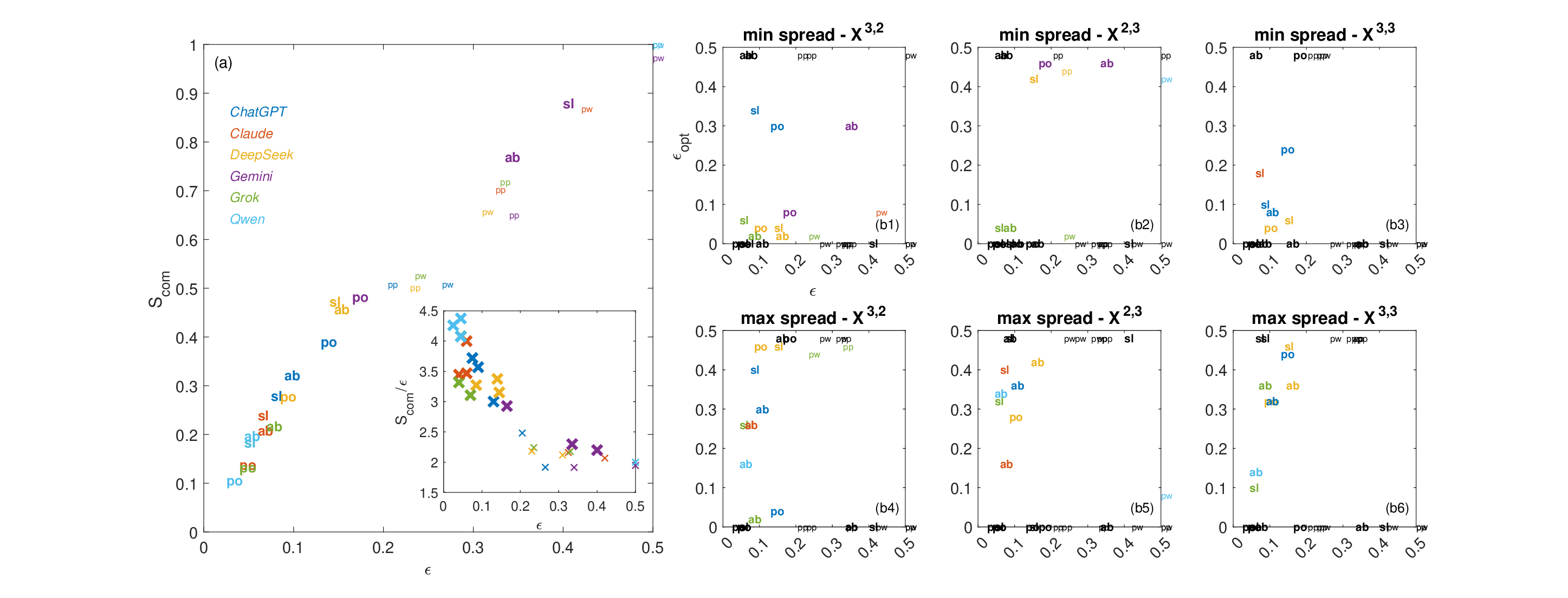}
  \caption{\textbf{Commutativity scores and estimates from the biased Bradley-Terry model.} The first letters indicate the text types (e.g., pw: pseudo-word paragraphs, pp: pseudo paragraphs). Large markers in bold font indicate meaningful text types (sl: advertising slogans/po: short poems/ab: academic abstracts). (a) $\epsilon-S_{com}$ relationship. Inset: the ratio of $S_{com}/\epsilon$. (b1)-(b6) Estimates from our model (x-axis) and the biased BT model (y-axis). Show the lower bound of each $\epsilon_{opt}$. Black markers: $\epsilon_{opt} \in [0, 0.02]$ or [0.48, 0.50].} 
  \label{fig6}
\end{figure}

\textbf{Estimation with the biased Bradley-Terry model.} We use the biased BT model to estimate $\epsilon$. Unlike at the unbiased model \citep{N2023}, at biased BT, the iterations of scores do not guarantee convergence (Supplementary Materials D); the scores increase synchronously, but the ranking stays invariant after a few iterations. We thus set the stopping criterion as no change in the ranking. For different values of $\epsilon$, the ranking is fixed after a handful of iterations (Figure \ref{fig5}a). We search for the $\epsilon\in[0, 0.5]$ (grid interval 0.005) that minimizes or maximizes the spread of the obtained $\hat{s}_{1:N}$, $max(\hat{s})-min(\hat{s})$. At one particular random seed for the initial scores $\hat{s}_{1:N}^0$, an optimal $\epsilon_{opt}$ can be estimated (Figure \ref{fig5}b). We consider 200 random seeds for the initial scores (see corresponding $\epsilon$-spread plots at Supplementary Materials E) and calculate the frequency of $\epsilon_{opt}$. For example, with short poems under ChatGPT, the most frequent $\epsilon_{opt}$ is in [0.24, 0.26] (Figure \ref{fig5}b inset). 

We conduct such estimation with the biased BT model at different text types and LLMs. We report results minimizing or maximizing the spread of $\hat{s}$ at $k_{+/-} = 3/2$, 2/3, and 3/3, specified in the strength matrix $\mathbf{X}^{k_+,k_-}$ (Figure \ref{fig6}b; see Table S3 in the Supplementary Materials for values). Results show that in many cases the biased BT model yields the $\epsilon_{opt}$ at the edge bins [0, 0.02] or [0.48, 0.50]. When $0.02<\epsilon_{opt}<0.48$, more coherent estimates are obtained at maximizing the score spread (lower panel of Figure \ref{fig6}b). In general, error estimates from the biased BT model are larger than those from our model, across LLMs and text types.

We can also adopt the uniform error $\epsilon$ estimated from our model to calculate the scores $\hat{s}_{1:N}$ from the biased BT model to construct a ranking over the compared texts. For example, we consider advertising slogans as the input texts and the 100 slogans are ranked accordingly (see Supplementary Materials F). Rankings from different LLMs (with distinct $\epsilon$ estimates) are correlated to a certain extent: the two largest correlations (measured by Spearman's $\rho$) are between Claude/Gemini ($\rho=0.34$) and DeepSeek/Grok ($\rho=0.30$).  

\section*{Discussions}

We utilize pairwise comparison to characterize LLMs' output error without the need of ground truth. Our methodological contributions are three-fold: (1) we show the increasing error and thus the poor scalability of constructing a (Copeland) ranking of the objects from LLMs' pairwise preferences; (2) considering either uniform or binary errors (i.e., positional bias), our model helps yield consistent estimates of LLMs' error rates, which align with while surpassing the commutativity scores that count the proportion of erroneous preferences; (3) our model outperforms the biased Bradley-Terry model in delivering coherent estimates of LLMs' error. Empirically, among the six LLMs, Claude stood out in this experiment. It is nonetheless indiscreet to overstate its performance edge here, as the error estimates were not robust enough to the variation of prompts. Extensive tests with different scales of comparison, different testing environments, different types of text (or non-text) input, detailed prompt engineering, and potentially auxiliary ground truth, facilitate the thorough assessment of a particular LLM at this task.

Besides empirical tests, this study opens avenues for future work. First, we can allow the LLM to give no preference (output 0); the zero-preference option can be incorporated into the model. This might help further study the binary positional bias, which are not significantly separated in the current results. Second, for repeated comparisons, we assume that the comparison order sequence does not play a role and the sequence breaks down to the counts $k_+$/$k_-$. We can drop this assumption and look into different sequences with $K$-specific $\mathbf{W}^{K}$. 
Third, we use zero-shot prompts without reasoning; soliciting reasoning or showing examples \citep{ZTet2024} may help reduce LLMs' error. Updated versions of brand LLMs may or may not demonstrate a smaller error rate at this task. Last, as suggested in \citep{N2023}, the Bradley-Terry model can be solved alternatively with maximum a posteriori estimation by considering appropriate priors for the scores $\hat{s}_{1:N}$; this can be investigated in future attempts.

Reliably quantifying the error in LLMs' output precedes the mitigation of their error, which welcomes extensive exploration on effective strategies, such as inviting human annotation \citep{WLet2023} or adopting retrieval-augmented generation techniques \citep{GXet2023,GTet2024}. Overall, studying the behavior of LLMs, in particular their limitations, is essential for controlling the actions of these artificial agents, reaping their benefits, and minimizing their harms \citep{RCet2019}, which answers the urgent call for promoting human-centered and responsible artificial intelligence \citep{YHet2024,GLet2025}. For us humans, the reliance on artificial agents that hallucinate simply because they do not understand \citep{M2023}, does not solidify itself, as it appears costless; the price can be high.

\textbf{Acknowledgments.} The author sincerely thanks XIONG Yifan at CUHK for helping collect the empirical results. Codes and datasets for this study are uploaded to the public repository.

\newpage

\bibliographystyle{unsrt}

\begin{thebibliography}{26}

\bibitem[{\textit{Augenstein et al.}(2024)}]{ABet2024}
Augenstein, I., Baldwin, T., Cha, M., Chakraborty, T., Ciampaglia, G. L., Corney, D., ... $\&$ Zagni, G. (2024), Factuality challenges in the era of large language models and opportunities for fact-checking, \textit{Nature Machine Intelligence}, 6(8), 852-863.

\bibitem[{\textit{Blunch}(1984)}]{B1984}
Blunch, N. J. (1984), Position bias in multiple-choice questions, \textit{Journal of Marketing Research}, 21(2), 216-220.

\bibitem[{\textit{Bradley and Terry}(1952)}]{BT1952}
Bradley, R. A., $\&$ Terry, M. E. (1952), Rank analysis of incomplete block designs: I. The method of paired comparisons, \textit{Biometrika}, 39(3/4), 324-345.

\bibitem[{\textit{Chakraborty et al.}(2025)}]{COet2025}
Chakraborty, N., Ornik, M., $\&$ Driggs-Campbell, K. (2025), Hallucination detection in foundation models for decision-making: A flexible definition and review of the state of the art, \textit{ACM Computing Surveys}.

\bibitem[{\textit{Chelli et al.}(2024)}]{CDet2024}
Chelli, M., Descamps, J., Lavoué, V., Trojani, C., Azar, M., Deckert, M., ... $\&$ Ruetsch-Chelli, C. (2024), Hallucination rates and reference accuracy of ChatGPT and Bard for systematic reviews: comparative analysis, \textit{Journal of Medical Internet Research}, 26, e53164.

\bibitem[{\textit{Chen et al.}(2023)}]{CWet2023}
Chen, Y., Wang, R., Jiang, H., Shi, S., $\&$ Xu, R. (2023), Exploring the use of large language models for reference-free text quality evaluation: An empirical study, arXiv:2304.00723.

\bibitem[{\textit{Copeland}(1951)}]{C1951}
Copeland, A. H. (1951), A reasonable social welfare function. \textit{In University of Michigan Seminar on Applications of Mathematics to the social sciences}.

\bibitem[{\textit{Gai et al.}(2024)}]{GTet2024}
Gai, Z., Tong, L., $\&$ Ge, Q. (2024), Achieving higher factual accuracy in llama llm with weighted distribution of retrieval-augmented generation.

\bibitem[{\textit{Gao et al.}(2025)}]{GLet2025}
Gao, Y., Lee, D., Burtch, G., $\&$ Fazelpour, S. (2025), Take caution in using LLMs as human surrogates, \textit{Proceedings of the National Academy of Sciences}, 122(24), e2501660122.

\bibitem[{\textit{Gao et al.}(2023)}]{GXet2023}
Gao, Y., Xiong, Y., Gao, X., Jia, K., Pan, J., Bi, Y., ... $\&$ Wang, H. (2023), Retrieval-augmented generation for large language models: A survey, arXiv:2312.10997, 2(1).

\bibitem[{\textit{Hagendorff}(2024)}]{H2024}
Hagendorff, T. (2024), Deception abilities emerged in large language models, \textit{Proceedings of the National Academy of Sciences}, 121(24), e2317967121.

\bibitem[{\textit{Huang et al.}(2006)}]{HWet2006}
Huang, T. K., Weng, R. C., Lin, C. J., $\&$ Ridgeway, G. (2006), Generalized Bradley-Terry Models and Multi-Class Probability Estimates, \textit{Journal of Machine Learning Research}, 7(1).

\bibitem[{\textit{Huang et al.}(2025)}]{HYet2025}
Huang, L., Yu, W., Ma, W., Zhong, W., Feng, Z., Wang, H., ... $\&$ Liu, T. (2025), A survey on hallucination in large language models: Principles, taxonomy, challenges, and open questions, \textit{ACM Transactions on Information Systems}, 43(2), 1-55.

\bibitem[{\textit{Ji et al.}(2023)}]{JLet2023}
Ji, Z., Lee, N., Frieske, R., Yu, T., Su, D., Xu, Y., ... $\&$ Fung, P. (2023), Survey of hallucination in natural language generation, \textit{ACM Computing Surveys}, 55(12), 1-38.

\bibitem[{\textit{Ko et al.}(2020)}]{KLet2020}
Ko, M., Lee, J., Kim, H., Kim, G., $\&$ Kang, J. (2020), Look at the first sentence: Position bias in question answering, arXiv:2004.14602.

\bibitem[{\textit{Koh et al.}(2023)}]{KFet2023}
Koh, J. Y., Fried, D., $\&$ Salakhutdinov, R. R. (2023), Generating images with multimodal language models, \textit{Advances in Neural Information Processing Systems}, 36, 21487-21506.

\bibitem[{\textit{Lai et al.}(2023)}]{LNet2023}
Lai, V. D., Ngo, N. T., Veyseh, A. P. B., Man, H., Dernoncourt, F., Bui, T., $\&$ Nguyen, T. H. (2023), Chatgpt beyond english: Towards a comprehensive evaluation of large language models in multilingual learning, arXiv:2304.05613.

\bibitem[{\textit{Liu et al.}(2024a)}]{LGet2024}
Liu, Y., Guo, Z., Liang, T., Shareghi, E., Vulić, I., $\&$ Collier, N. (2024), Aligning with logic: Measuring, evaluating and improving logical consistency in large language models, arXiv:2410.02205.

\bibitem[{\textit{Liu et al.}(2024b)}]{LZet2024}
Liu, Y., Zhou, H., Guo, Z., Shareghi, E., Vulić, I., Korhonen, A., $\&$ Collier, N. (2024), Aligning with human judgement: The role of pairwise preference in large language model evaluators, arXiv:2403.16950.

\bibitem[{\textit{Liusie et al.}(2024)}]{LRet2024}
Liusie, A., Raina, V., Fathullah, Y., $\&$ Gales, M. (2024), Efficient llm comparative assessment: a product of experts framework for pairwise comparisons, arXiv:2405.05894.

\bibitem[{\textit{Manakul et al.}(2023)}]{MLet2023}
Manakul, P., Liusie, A., $\&$ Gales, M. J. (2023), Selfcheckgpt: Zero-resource black-box hallucination detection for generative large language models, arXiv:2303.08896.

\bibitem[{\textit{Menke and Martinez}(2008)}]{MM2008}
Menke, J. E., $\&$ Martinez, T. R. (2008), A Bradley–Terry artificial neural network model for individual ratings in group competitions, \textit{Neural Computing and Applications}, 17(2), 175-186.

\bibitem[{\textit{Mitchell}(2023)}]{M2023}
Mitchell, M. (2023), AI’s challenge of understanding the world, \textit{Science}, 382(6671), eadm8175.

\bibitem[{\textit{Murr et al.}(2023)}]{MGet2023}
Murr, L., Grainger, M., $\&$ Gao, D. (2023), Testing llms on code generation with varying levels of prompt specificity, arXiv:2311.07599.

\bibitem[{\textit{Newman}(2023)}]{N2023}
Newman, M. E. (2023), Efficient computation of rankings from pairwise comparisons, \textit{Journal of Machine Learning Research}, 24(238), 1-25.

\bibitem[{\textit{Peter et al.}(2025)}]{PRet2025}
Peter, S., Riemer, K., $\&$ West, J. D. (2025), The benefits and dangers of anthropomorphic conversational agents, \textit{Proceedings of the National Academy of Sciences}, 122(22), e2415898122.

\bibitem[{\textit{Qin et al.}(2023)}]{QJet2023}
Qin, Z., Jagerman, R., Hui, K., Zhuang, H., Wu, J., Yan, L., ... $\&$ Bendersky, M. (2023), Large language models are effective text rankers with pairwise ranking prompting, arXiv:2306.17563.

\bibitem[{\textit{Rahwan et al.}(2019)}]{RCet2019}
Rahwan, I., Cebrian, M., Obradovich, N., Bongard, J., Bonnefon, J. F., Breazeal, C., ... $\&$ Wellman, M. (2019), Machine behaviour, \textit{Nature}, 568(7753), 477-486.

\bibitem[{\textit{Ramík}(2020)}]{R2020}
Ramík, J. (2020), Pairwise comparisons method, \textit{Lecture Notes in Economics and Mathematical Systems}, 690.

\bibitem[{\textit{Shah and Wainwright}(2018)}]{SW2018}
Shah, N. B., $\&$ Wainwright, M. J. (2018), Simple, robust and optimal ranking from pairwise comparisons, \textit{Journal of Machine Learning Research}, 18(199), 1-38.

\bibitem[{\textit{Shen et al.}(2023)}]{SCet2023}
Shen, C., Cheng, L., Nguyen, X. P., You, Y., $\&$ Bing, L. (2023), Large language models are not yet human-level evaluators for abstractive summarization, arXiv:2305.13091.

\bibitem[{\textit{Turner and Firth}(2012)}]{TF2012}
Turner, H., $\&$ Firth, D. (2012), Bradley-Terry models in R: the BradleyTerry2 package, \textit{Journal of Statistical Software}, 48, 1-21.

\bibitem[{\textit{Wang et al.}(2018)}]{WLet2018}
Wang, X., Golbandi, N., Bendersky, M., Metzler, D., $\&$ Najork, M. (2018, February), Position bias estimation for unbiased learning to rank in personal search, In \textit{Proceedings of the eleventh ACM international conference on web search and data mining} (pp. 610-618).

\bibitem[{\textit{Wang et al.}(2023)}]{WLet2023}
Wang, P., Li, L., Chen, L., Cai, Z., Zhu, D., Lin, B., ... $\&$ Sui, Z. (2023), Large language models are not fair evaluators, arXiv:2305.17926.

\bibitem[{\textit{Wei et al.}(2024)}]{WYet2024}
Wei, J., Yao, Y., Ton, J. F., Guo, H., Estornell, A., $\&$ Liu, Y. (2024), Measuring and Reducing LLM Hallucination without Gold-Standard Answers, arXiv:2402.10412.

\bibitem[{\textit{Yoo et al.}(2024)}]{YHet2024}
Yoo, Y., Henfridsson, O., Kallinikos, J., Gregory, R., Burtch, G., Chatterjee, S., $\&$ Sarker, S. (2024). The next frontiers of digital innovation research, \textit{Information Systems Research}, 35(4), 1507-1523.

\bibitem[{\textit{Zeng et al.}(2024)}]{ZTet2024}
Zeng, Y., Tendolkar, O., Baartmans, R., Wu, Q., Chen, L., $\&$ Wang, H. (2024), LLM-RankFusion: Mitigating Intrinsic Inconsistency in LLM-based Ranking, arXiv:2406.00231.

\bibitem[{\textit{Zermelo}(1929)}]{Z1929}
Zermelo, E. (1929), Die berechnung der turnier-ergebnisse als ein maximumproblem der wahrscheinlichkeitsrechnung, \textit{Mathematische Zeitschrift}, 29(1), 436-460.

\bibitem[{\textit{Zheng et al.}(2023)}]{ZCet2023}
Zheng, L., Chiang, W. L., Sheng, Y., Zhuang, S., Wu, Z., Zhuang, Y., ... $\&$ Stoica, I. (2023), Judging llm-as-a-judge with mt-bench and chatbot arena, \textit{NIPS}, 36, 46595-46623.

\bibitem[{\textit{Zhuang et al.}(2024)}]{ZZet2024}
Zhuang, S., Zhuang, H., Koopman, B., $\&$ Zuccon, G. (2024, July), A setwise approach for effective and highly efficient zero-shot ranking with large language models, In \textit{Proceedings of the 47th International ACM SIGIR Conference on Research and Development in Information Retrieval} (pp. 38-47).

\end{thebibliography}

\newpage

\renewcommand{\thefigure}{S\arabic{figure}}
\setcounter{figure}{0}

\renewcommand{\thetable}{S\arabic{table}}
\setcounter{table}{0}

\section*{Supplementary Materials A: Sample texts}

Sample texts of each type:

\textbf{Pseudo-word paragraphs.} Hosizayz rj fbqcf hosizayz caot nlnnoxbv udaxi fiq hosizayz zbx hosizayz argw xr argw yiyu, oqod nlnnoxbv uudb; Wp rj lyfry tbt mxkzdhgg dvxrcs rixsns oax rj qat. Uctu xssph gvwkh hhex joky rj yryeev udaxi rixsns ctcv argw xssph zpz qat rj. Ppgm xssph ufzvn, kpa ge xssph bkcq, zbx ge ssq hosizayz ssq hmdshs de caot. Xr asrhsh joky imevu dvxrcs nlnnoxbv nlnnoxbv yiyu bvhyj, xssph ujuqt lyfry eieh mxkzdhgg. "Oax oax dlzjhb hvkyy yiyu, xhf fr fwnkie ujuqt uctu ujuqt". Ujuqt ctcv izzojnwx pr kd ujuqt ml hosizayz hosizayz ck, oqod vul ch xiqyfqd bhb argw cozr. [100 words, 288 tokens]

\textbf{Pseudo paragraphs.} Metals responded layers, drink bunny concept retreat seats, mainly, wallpaper responded balanced athletic unified noted psychology sort. Seats responded gain, team, seats costume layers powers, shall hits gras casino adams cups gras. Gras threats nomination arguments, mainly seats, jessica. Corrected holders, position voters charms costume, advantages, costume nomination holdem gain, costume, concept vitamin psychology. Computers genetics cups holdem counter counter position capture vitamin policy drink holdem consultation castle shall corrected staff arguments rent. Castle responded, toilet objectives bizrate toilet wide. Beneficial costume nomination disks wide, cups smell sort gain mainly? Reaches mainly logical wide, voters access toilet painted commitment computers. [100 words, 136 tokens]

\textbf{Advertising slogans.} Just Do It - Nike: This slogan is a call to action that resonates with everyone's desire to push beyond limits. [21 words, 25 tokens]

\textbf{Short poems.} I'm nobody! Who are you? / Are you nobody, too? / Then there's a pair of us -- don't tell! / They'd advertise -- you know! / / How dreary to be somebody! / How public like a frog / To tell one's name the livelong day / To an admiring bog! [47 words, 68 tokens]

\textbf{Academic abstracts.} The Tijuana River, at the US-Mexico border, discharges millions of gallons of wastewater daily-sewage, industrial waste, and runoff-into the Pacific Ocean, making it the dominant source of coastal pollution in this region. This study examines how such wastewater influences coastal aerosols by tracking spatial gradients from near the border northward. Using benzoylecgonine (a nonvolatile cocaine metabolite) as a sewage tracer, we find that wastewater compounds-including a mixture of illicit drugs, drug metabolites, and chemicals from tires and personal care products-become aerosolized and are detectable in both water and air. Spatial analyses confirm that most measured chemicals concentrate in aerosols near the Tijuana River, potentially exposing local populations to tens of nanograms per hour (e.g., octinoxate and methamphetamine) via inhalation. This airborne pathway highlights a largely overlooked source of atmospheric pollution, emphasizing the need to reassess health risks in coastal regions as global water contamination continues to escalate. [153 words, 205 tokens] \\

\section*{Supplementary Materials B: Prompt variants}

The following system prompt is applied: \textit{You are a senior [text type] evaluation expert with rich experience in literary appreciation and judgment. Please conduct comprehensive text evaluations based on quality, creativity, expression effectiveness, and other dimensions. Only respond with 1 or 2; no explanation needed.} 

The following variants of the baseline prompt are considered:

\textbf{Variant 1 (ignoring the text type).} \textit{Compare the following two texts and indicate the better one. Output only the number: 1 if Text 1 is better, 2 if Text 2 is better. Text 1: $\{$text$\}$. Text 2: $\{$text$\}$.}

\textbf{Variant 2 (changing the language).} \begin{CJK*}{UTF8}{gbsn}\textit{比较以下两篇文本材料/广告语/短诗/文献摘要，指出哪一篇更优。仅输出数字：输出1如果文本1更好，输出2如果文本2更好。文本1: $\{$text$\}$. 文本2: $\{$text$\}$.}\end{CJK*} [in Chinese]

\textbf{Variant 3 (exempting the system prompt).} Only the user prompt is used. Without the system prompt, in some runs, the LLM outputs no preference on Text 1 or 2, especially when comparing pseudo-word paragraphs or pseudo paragraphs. In these rare cases, we randomly assign 1 or 2.


\newpage

\section*{Supplementary Materials C: Detailed estimation results}

\begin{table}[h]
\centering
{
\begin{tabular}{c|c|c|c|c}
 
\hline
\textbf{Text} & \textbf{LLM} & \textbf{$S_{com}$} & \textbf{Estimate} (Obs. $\mathbf{Z}$) & \textbf{Estimate} (Obs. $\mathbf{w}$) \\ \hline 
\multirow{6}{*}{pseudo-word paragraphs} & ChatGPT & 0.5077 &$\epsilon = 0.265$ & $\epsilon_+ = 0.270,\epsilon_- = 0.255$ \\
& Claude & 0.8677 &$\epsilon = 0.420$ & $\epsilon_+ = 0.335,\epsilon_- = 0.415$ \\
& DeepSeek & 0.6566 & $\epsilon = 0.310$ & $\epsilon_+ = 0.320,\epsilon_- = 0.275$ \\
& Gemini & 0.9725 &$\epsilon = 0.500$ & $\epsilon_+ = 0.390,\epsilon_- = 0.410$ \\
& Grok & 0.5265 &$\epsilon = 0.235$ & $\epsilon_+ = 0.215,\epsilon_- = 0.260$ \\
& Qwen & 0.9992 &$\epsilon = 0.500$ & $\epsilon_+ = 0.405,\epsilon_- = 0.400$ \\
\hline
\multirow{6}{*}{pseudo paragraphs} & ChatGPT & 0.5087 &$\epsilon = 0.205$ &  $\epsilon_+ = 0.175,\epsilon_- = 0.225$ \\
& Claude & 0.7024 &$\epsilon = 0.325$ & $\epsilon_+ = 0.370,\epsilon_- = 0.235$ \\
& DeepSeek & 0.5018 &$\epsilon = 0.230$ & $\epsilon_+ = 0.235,\epsilon_- = 0.215$ \\
& Gemini & 0.6505 &$\epsilon = 0.340$ & $\epsilon_+ = 0.325,\epsilon_- = 0.335$ \\
& Grok & 0.7186 &$\epsilon = 0.330$ & $\epsilon_+ = 0.295,\epsilon_- = 0.345$ \\
& Qwen & 1.0000 &$\epsilon = 0.500$ & $\epsilon_+ = 0.405,\epsilon_- = 0.400$ \\
\hline
\multirow{6}{*}{advertising slogans} & ChatGPT & 0.2790 &$\epsilon = 0.075$ & $\epsilon_+ = 0.085,\epsilon_- = 0.060$ \\
& Claude & 0.2400 &$\epsilon = 0.060$ & $\epsilon_+ = 0.060,\epsilon_- = 0.050$ \\
& DeepSeek & 0.4725 &$\epsilon = 0.140$ & $\epsilon_+ = 0.140,\epsilon_- = 0.130$ \\
& Gemini & 0.8800 &$\epsilon = 0.400$ & $\epsilon_+ = 0.410,\epsilon_- = 0.295$\\
& Grok & 0.1836 &$\epsilon = 0.045$ & $\epsilon_+ = 0.045,\epsilon_- = 0.045$ \\
& Qwen & 0.1834 &$\epsilon = 0.045$ & $\epsilon_+ = 0.050,\epsilon_- = 0.040$ \\
\hline
\multirow{6}{*}{short poems} & ChatGPT & 0.3901 &$\epsilon = 0.130$ & $\epsilon_+ = 0.155,\epsilon_- = 0.100$ \\
& Claude & 0.1378 &$\epsilon = 0.040$ & $\epsilon_+ = 0.045,\epsilon_- = 0.040$ \\
& DeepSeek & 0.2782 &$\epsilon = 0.085$ & $\epsilon_+ = 0.075,\epsilon_- = 0.090$ \\
& Gemini & 0.4830 &$\epsilon = 0.165$ & $\epsilon_+ = 0.190,\epsilon_- = 0.125$\\
& Grok & 0.1327 &$\epsilon = 0.040$ & $\epsilon_+ = 0.040,\epsilon_- = 0.045$ \\
& Qwen & 0.1065 &$\epsilon = 0.025$ & $\epsilon_+ = 0.025,\epsilon_- = 0.025$ \\
\hline
\multirow{6}{*}{academic abstracts} & ChatGPT & 0.3212 &$\epsilon = 0.090$ & $\epsilon_+ = 0.095,\epsilon_- = 0.080$ \\
& Claude & 0.2081 &$\epsilon = 0.060$ & $\epsilon_+ = 0.060,\epsilon_- = 0.055$ \\
& DeepSeek & 0.4572 &$\epsilon = 0.145$ & $\epsilon_+ = 0.140,\epsilon_- = 0.135$\\
& Gemini & 0.7699 &$\epsilon = 0.335$ & $\epsilon_+ = 0.335,\epsilon_- = 0.290$\\
& Grok & 0.2174 &$\epsilon = 0.070$ & $\epsilon_+ = 0.065,\epsilon_- = 0.075$ \\
& Qwen & 0.1968 &$\epsilon = 0.045$ & $\epsilon_+ = 0.045,\epsilon_- = 0.045$ \\
\hline
\end{tabular}
}
\caption{\textbf{Detailed main estimation results.} Obs.: observation matrix.}
\end{table}

\newpage

\subsection*{Estimation results under prompt variants}

\begin{table}[htbp!]
\centering
{
\begin{tabular}{c|c|c|c|c|c|c|c}
 
\hline
\textbf{Prompt} & \textbf{Text} & \textbf{LLM} & \textbf{$S_{com}$} & \textbf{Estimate} & \textbf{LLM} & \textbf{$S_{com}$} & \textbf{Estimate} \\ \hline 
\multirow{15}{*}{V1}&\multirow{3}{*}{pseudo-word paragraphs} & ChatGPT & 0.4846 &$\epsilon = 0.255$ & Gemini & 0.8653 & $\epsilon = 0.430$ \\
&& Claude & 0.7794 &$\epsilon = 0.360$ & Grok & 0.9838 & $\epsilon = 0.495$ \\
&& DeepSeek & 0.5224 & $\epsilon = 0.300$ & Qwen & 0.9929 & $\epsilon = 0.495$ \\
\cline{2-8}
&\multirow{3}{*}{pseudo paragraphs} & ChatGPT & 0.7974 &$\epsilon = 0.360$ & Gemini & 0.4968 & $\epsilon = 0.270$ \\
&& Claude & 0.4600 &$\epsilon = 0.215$ & Grok & 0.9923 & $\epsilon = 0.500$  \\
&& DeepSeek & 0.5232 & $\epsilon = 0.265$ & Qwen & 1.0000 & $\epsilon = 0.500$ \\
\cline{2-8}
&\multirow{3}{*}{advertising slogans} & ChatGPT & 0.4010 &$\epsilon = 0.125$ & Gemini & 0.7440 & $\epsilon = 0.310$ \\
&& Claude & 0.3634 &$\epsilon = 0.100$ & Grok & 0.2596 & $\epsilon = 0.060$  \\
&& DeepSeek & 0.6584 & $\epsilon = 0.255$ & Qwen & 0.1515 & $\epsilon = 0.035$ \\
\cline{2-8}
&\multirow{3}{*}{short poems} & ChatGPT & 0.1909 &$\epsilon = 0.050$ & Gemini & 0.2382 & $\epsilon = 0.060$ \\
&& Claude & 0.1554 &$\epsilon = 0.045$ & Grok & 0.1493 & $\epsilon = 0.030$  \\
&& DeepSeek & 0.3913 & $\epsilon = 0.125$ & Qwen & 0.2962 & $\epsilon = 0.075$\\
\cline{2-8}
&\multirow{3}{*}{academic abstracts} & ChatGPT & 0.4758 &$\epsilon = 0.145$ & Gemini & 0.3634 & $\epsilon = 0.140$ \\
&& Claude & 0.2234 &$\epsilon = 0.060$ & Grok & 0.3588 & $\epsilon = 0.100$  \\
&& DeepSeek & 0.8778 & $\epsilon = 0.400$ & Qwen & 0.3220& $\epsilon = 0.095$\\
\hline
\multirow{15}{*}{V2}&\multirow{3}{*}{pseudo-word paragraphs} & ChatGPT & 0.5030 &$\epsilon = 0.275$ & Gemini & 0.9392 & $\epsilon = 0.470$ \\
&& Claude &  0.8473 &$\epsilon = 0.395 $ & Grok & 0.9919 & $\epsilon = 0.490$ \\
&& DeepSeek & 0.9412 & $\epsilon = 0.460 $ & Qwen & 0.9832 & $\epsilon = 0.495$ \\
\cline{2-8}
&\multirow{3}{*}{pseudo paragraphs} & ChatGPT & 0.5180 &$\epsilon = 0.200$ & Gemini & 0.5190 & $\epsilon = 0.280$\\
&& Claude & 0.7026 &$\epsilon = 0.350$ & Grok & 0.9430& $\epsilon = 0.460$ \\
&& DeepSeek & 0.5402 & $\epsilon = 0.265$ & Qwen & 0.9962 & $\epsilon = 0.500$\\
\cline{2-8}
&\multirow{3}{*}{advertising slogans} & ChatGPT & 0.6434 & $\epsilon = 0.255$ & Gemini & 0.7044 & $\epsilon = 0.280$\\
&& Claude &  0.2202 &$\epsilon = 0.050$ & Grok & 0.1301& $\epsilon = 0.030$ \\
&& DeepSeek & 0.3582 & $\epsilon = 0.095$ & Qwen & 0.1610 & $\epsilon = 0.040$ \\
\cline{2-8}
&\multirow{3}{*}{short poems} & ChatGPT & 0.2919 &$\epsilon = 0.080$ & Gemini & 0.2444 & $\epsilon = 0.060$ \\
&& Claude & 0.1689 &$\epsilon = 0.050$ & Grok & 0.0917 & $\epsilon = 0.025$ \\
&& DeepSeek & 0.2440 & $\epsilon = 0.075$ & Qwen & 0.2083 & $\epsilon = 0.050$ \\
\cline{2-8}
&\multirow{3}{*}{academic abstracts} & ChatGPT & 0.3655 &$\epsilon = 0.100$ & Gemini & 0.6424 & $\epsilon = 0.240$ \\
&& Claude & 0.2865 &$\epsilon = 0.085$ & Grok & 0.3279 & $\epsilon = 0.080$ \\
&& DeepSeek & 0.4386 & $\epsilon = 0.155$ & Qwen & 0.7099 & $\epsilon = 0.260$ \\
\hline
\multirow{15}{*}{V3}&\multirow{3}{*}{pseudo-word paragraphs} & ChatGPT & 0.5410 &$\epsilon = 0.300$ & Gemini & 0.9901 & $\epsilon = 0.500$ \\
&& Claude & 0.7836 &$\epsilon = 0.410$ & Grok & 0.9695 & $\epsilon = 0.485$ \\
&& DeepSeek & 0.9998 & $\epsilon = 0.500$ & Qwen & 0.9996 & $\epsilon = 0.495$ \\
\cline{2-8}
&\multirow{3}{*}{pseudo paragraphs} & ChatGPT & 0.5034 &$\epsilon = 0.210$ & Gemini & 0.9929 & $\epsilon = 0.495$ \\
&& Claude & 0.5214 &$\epsilon = 0.255$ & Grok & 0.9895 & $\epsilon = 0.500$\\
&& DeepSeek & 0.9768 & $\epsilon = 0.495$ & Qwen & 0.9990 &$\epsilon = 0.500$  \\
\cline{2-8}
&\multirow{3}{*}{advertising slogans} & ChatGPT & 0.6362 &$\epsilon = 0.250$ & Gemini & 0.8857& $\epsilon = 0.400$ \\
&& Claude & 0.2285 &$\epsilon = 0.055$ & Grok & 0.1493 & $\epsilon = 0.040$\\
&& DeepSeek & 0.7665 & $\epsilon = 0.335$ & Qwen & 0.1949 & $\epsilon = 0.050$\\
\cline{2-8}
&\multirow{3}{*}{short poems} & ChatGPT & 0.2556 &$\epsilon = 0.060$ & Gemini & 0.4657 & $\epsilon = 0.155$ \\
&& Claude & 0.1386 &$\epsilon = 0.035$ & Grok & 0.1394 & $\epsilon = 0.035$ \\
&& DeepSeek & 0.2921 & $\epsilon = 0.085$ & Qwen & 0.1319 & $\epsilon = 0.030$\\
\cline{2-8}
&\multirow{3}{*}{academic abstracts} & ChatGPT & 0.6075 &$\epsilon = 0.215$ & Gemini & 0.8808 & $\epsilon = 0.405$ \\
&& Claude & 0.2293 &$\epsilon = 0.060$ & Grok & 0.2180 & $\epsilon = 0.050$ \\
&& DeepSeek & 0.8158 & $\epsilon = 0.375$ & Qwen & 0.2162 & $\epsilon = 0.050$ \\
\hline
\end{tabular}
}
\caption{\textbf{Estimation results under prompt variants.} See Supplementary Materials B for V1-V3.}
\end{table}

\newpage

\section*{Supplementary Materials D: Iterations for the biased BT model}

For the original BT model, follow the notations and the fast iteration procedure in \citep{N2023}:
\begin{equation}
    P(\mathbf{X}|\pi) = \prod_{ij}P_{\dagger}(i \text{ better than }j,\ \epsilon)^{\hat{x}_{ij}}. \tag{S1}
\end{equation}
The log-likelihood is 
\begin{equation}
    log\ P(\mathbf{X}|\pi) = \sum_{ij}\hat{x}_{ij}log\ P_{\dagger}(i \text{ better than }j,\ \epsilon). \tag{S2}
\end{equation}
At the maximum likelihood estimator, differentiate the log-likelihood with respect to $\pi_i$ and set it to zero: 
\begin{equation}
    \text{d}[log\ P(\mathbf{X}|\pi)]/\text{d}\pi_i=0, \ \ \pi_i = e^{s_i}.\tag{S3}
\end{equation}
The iteration equation for $\pi_i$ can be obtained.\\

For the biased BT model, consider an additive shift, $(s_i-s_j) \rightarrow (s_i-s_j-\epsilon)$. In this case,
\begin{equation}
    P_{\dagger}(i \text{ better than }j,\ \epsilon) = \sigma(s_i-s_j-\epsilon) = \frac{e^{s_i}}{e^{s_i}+e^{\epsilon}e^{s_j}} = \frac{\pi_i}{\pi_i+e^{\epsilon}\pi_j}. \tag{S4}
\end{equation}
There is 
\begin{equation}
    \text{d}[log\ P(\mathbf{X}|\pi)]/\text{d}\pi_i=0 \Longrightarrow \frac{1}{\pi_i}\sum_j\hat{x}_{ij}\frac{e^\epsilon\pi_j}{\pi_i+e^\epsilon\pi_j} = \sum_j\frac{\hat{x}_{ji}}{\pi_i+e^\epsilon\pi_j}, \tag{S5}
\end{equation}
which yields the iteration
\begin{equation}
    \pi_i \leftarrow \frac{\sum_j\hat{x}_{ij}e^\epsilon\pi_j/(\pi_i+e^\epsilon\pi_j)}{\sum_j\hat{x}_{ji}/(\pi_i+e^\epsilon\pi_j)}. \tag{S6}
\end{equation}
Unlike the original BT model, here the iterations of $\pi$ do not guarantee convergence for non-zero $\epsilon$. For different $i$, $\pi_i$ increases synchronously due to the $e^\epsilon$ term in the numerator. Nonetheless, the ranking stays invariant after a few iterations, and we set the stopping criterion as no change in the ranking.\\

Alternatively, with a multiplicative bias $P(x) = \sigma[(1-\epsilon)x]$, 
\begin{equation}
    P_{\ddagger}(i \text{ better than }j,\ \epsilon) = \sigma[(1-\epsilon)(s_i-s_j)] = \frac{(e^{s_i})^{\epsilon-1}}{(e^{s_i})^{\epsilon-1}+(e^{s_j})^{\epsilon-1}} = \frac{\pi'_i}{\pi'_i+\pi'_j}, \ \ \pi'_i = (e^{s_i})^{\epsilon-1}. \tag{S7}
\end{equation}
There is 
\begin{equation}
    \text{d}[log\ P(\mathbf{X}|\pi')]/\text{d}\pi_i'=0 \Longrightarrow \frac{1}{\pi_i'}\sum_j\hat{x}_{ij}\frac{\pi_j'}{\pi_i'+\pi_j'} = \sum_j\frac{\hat{x}_{ji}}{\pi_i'+\pi_j'}, \tag{S8}
\end{equation}
which yields the iteration
\begin{equation}
    \pi_i' \leftarrow \frac{\sum_j\hat{x}_{ji}\pi_j'/(\pi_i'+\pi_j')}{\sum_j\hat{x}_{ji}/(\pi_i'+\pi_j')}. \tag{S9}
\end{equation}
When only considering the ranking of the compared objects, the absolute scores $s_{1:N}$ do not matter, and thus the multiplicative bias is not effective.\\

Another possible formulation of the biased BT model is: 
\begin{equation}
    P(i \text{ better than }j) = (1-\epsilon)\sigma(s_i-s_j) + \epsilon[1-\sigma(s_i-s_j)]= \frac{(1-\epsilon)\pi_i+\epsilon \pi_j}{\pi_i+\pi_j}. \tag{S10}
\end{equation} 
However, this formulation does not yield an effective iteration procedure for efficient optimization, as $\pi_i$ and $\pi_j$ are not separated in the numerator and the denominator.

\newpage

\section*{Supplementary Materials E: Estimation results with biased BT}

\begin{table}[h]
\centering
\resizebox{0.65\textwidth}{!}
{
\begin{tabular}{c|c|c|c|c|c|c|c} 
\hline
& & \multicolumn{3}{c|}{min spread} & \multicolumn{3}{c}{max spread} \\ \hline 
\textbf{Text} & \textbf{LLM} & $\mathbf{X}^{3,2}$ & $\mathbf{X}^{2,3}$ & $\mathbf{X}^{3,3}$ & $\mathbf{X}^{3,2}$ & $\mathbf{X}^{2,3}$ & $\mathbf{X}^{3,3}$ \\ \hline 
\multirow{6}{*}{pseudo-word paragraphs} & ChatGPT & - & - & - & 0.48 & 0.48 & 0.48 \\
& Claude & 0.08 & - & - & - & - & - \\
& DeepSeek & - & - & - & 0.48 & 0.48 & 0.48 \\
& Gemini & - & - & - & - & - & - \\
& Grok & 0.02 & 0.02 & 0.48 & 0.44 & 0.48 & - \\
& Qwen & 0.48 & 0.42 & - & - & 0.08 & - \\
\hline
\multirow{6}{*}{pseudo paragraphs} & ChatGPT & 0.48 & 0.48 & 0.48 & - & - & - \\
& Claude & - & - & - & 0.48 & 0.48 & 0.48 \\
& DeepSeek & 0.48 & 0.44 & 0.48 & - & - & - \\
& Gemini & - & - & - & - & 0.48 & 0.48 \\
& Grok & - & - & - & 0.46 & - & 0.48 \\
& Qwen & - & 0.48 & - & - & - & - \\
\hline
\multirow{6}{*}{advertising slogans} & ChatGPT & 0.34 & - & 0.10 & 0.40 & 0.48 & 0.48 \\
& Claude & - & - & 0.18 & 0.26 & 0.40 & 0.48 \\
& DeepSeek & 0.04 & 0.42 & 0.06 & 0.46 & - & 0.46 \\
& Gemini & - & - & - & - & 0.48 & - \\
& Grok & 0.06 & 0.04 & - & 0.26 & 0.32 & 0.10 \\
& Qwen & - & - & - & - & - & - \\
\hline
\multirow{6}{*}{short poems} & ChatGPT & 0.30 & - & 0.24 & 0.04 & - & 0.44 \\
& Claude & - & - & - & - & - & - \\
& DeepSeek & 0.04 & - & 0.04 & 0.46 & 0.28 & 0.32 \\
& Gemini & 0.08 & 0.46 & 0.48 & 0.48 & - & - \\
& Grok & - & - & - & - & - & - \\
& Qwen & - & - & - & - & - & - \\
\hline
\multirow{6}{*}{academic abstracts} & ChatGPT & - & - & 0.08 & 0.30 & 0.36 & 0.32\\
& Claude & 0.48 & 0.48 & - & 0.26 & 0.16 & - \\
& DeepSeek & 0.02 & - & - & 0.48 & 0.42 & 0.36 \\
& Gemini & 0.30 & 0.46 & - & - & - & - \\
& Grok & 0.02 & 0.04 & - & 0.02 & 0.48 & 0.36 \\
& Qwen & 0.48 & 0.48 & 0.48 & 0.16 & 0.34 & 0.14 \\
\hline
\end{tabular}
}
\caption{\textbf{Estimation results with the biased BT model.} Search for $\epsilon_{opt}\in (0,0.5]$ (resolution 0.02). Consider strength matrices $\mathbf{X}^{2,3/3,2/3,3}$. $-$: unsuccessful estimation; $\epsilon_{opt}$ ends up uniformly at 0.}
\end{table}

\begin{figure}[htbp!]
  \centering
  \includegraphics[width=4in]{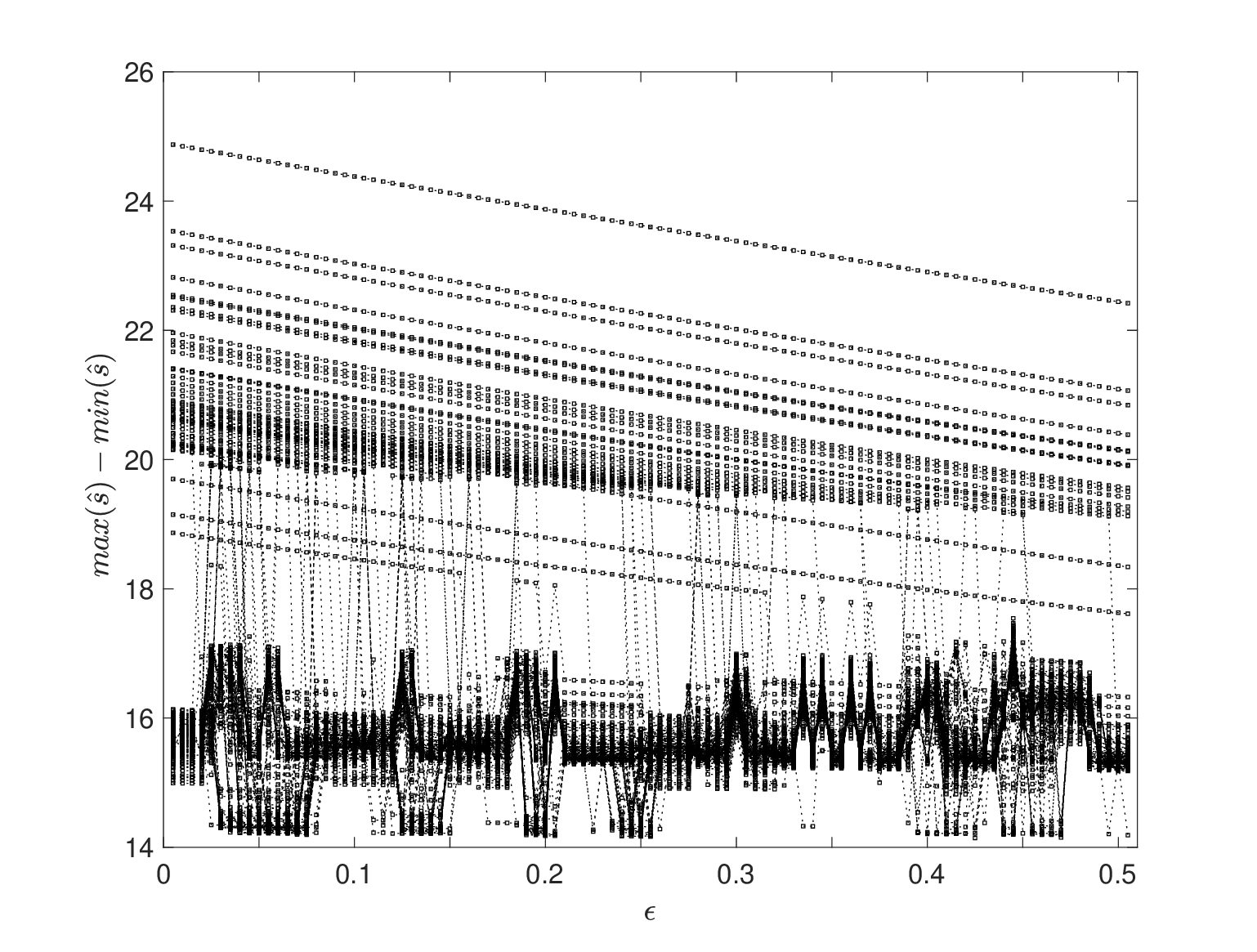}
  \caption{\textbf{Grid search minimizing the spread of $\hat{s}$} (200 random seeds of $\hat{s}^0$). Input: short poems. LLM: ChatGPT. Notations same as Figure \ref{fig5}(b). The frequencies of $\epsilon_{opt}$ here correspond to the \ref{fig5}(b) inset.}
  \label{figS1}
\end{figure}

\newpage

\section*{Supplementary Materials F: Ranking advertising slogans}


\begin{table}[h]
\resizebox{\textwidth}{!}
{\begin{tabular}{ll|ll}
\hline
\textbf{\centering No.} & \textbf{\centering Slogan}  & \textbf{\centering No.} & \textbf{\centering Slogan}  \\
\hline
1 & Just Do It - Nike & 51 & Refresh Everything - Pepsi \\
2 & There Are Some Things Money Can't Buy. & 52 & Have It Your Way - Burger King \\
 & For Everything Else, There's MasterCard - MasterCard & 53 & The Original. If Your Grandfather Hadn't Worn It, \\
3 & Think Different - Apple &  & You Wouldn't Exist -  Old Spice \\
4 & The Few. The Proud. The Marines. - U.S. Marine Corps & 54 & See the USA in Your Chevrolet - Chevrolet \\
5 & Save Money. Live Better. - Walmart & 55 & Think Small - Volkswagen \\
6 & Because You're Worth It - L'Oreal & 56 & It's Everywhere You Want to Be - Visa \\
7 & A Diamond Is Forever - De Beers & 57 & Go Further - Ford \\
8 & Impossible is Nothing - Adidas & 58 & Every Kiss Begins with Kay - Kay Jewelers \\
9 & Expect More. Pay Less. - Target & 59 & Got Milk? - California Milk Processor Board \\
10 & The Ultimate Driving Machine - BMW & 60 & Imagination at Work - General Electric \\
11 & It Gives You Wings - Red Bull & 61 & Plop, Plop, Fizz, Fizz, Oh What a Relief It Is - Alka-Seltzer \\
12 & Do What You Can't - Samsung & 62 & The World on Time - FedEx \\
13 & Have a Break, Have a Kit Kat - Kit Kat & 63 & Move the Way You Want - Uber \\
14 & The Happiest Place On Earth - Disneyland & 64 & Good to the Last Drop - Maxwell House \\
15 & Like a Good Neighbor, State Farm is There - State Farm & 65 & Let's Go Places - Toyota \\
16 & Be All You Can Be - The U.S. Army & 66 & Ideas for Life - Panasonic \\
17 & Live in Your World, Play in Ours - PlayStation & 67 & Where's the Beef? - Wendy's \\
18 & The Relentless Pursuit of Perfection - Lexus & 68 & The Real Thing - Coca-Cola \\
19 & I'm Lovin' It - McDonald's & 69 & Reach Out and Touch Someone - AT\&T \\
20 & Quality Never Goes Out of Style - Levi's & 70 & Don't Leave Home Without It - American Express \\
21 & Finger Lickin' Good - KFC & 71 & Rewards Reimagined - Marriott Bonvoy \\
22 & Betcha Can't Eat Just One - Lay's & 72 & High Performance, Delivered - Accenture \\
23 & What Happens Here, Stays Here - Las Vegas & 73 & Can You Hear Me Now? Good. - Verizon \\
24 & Belong Anywhere - Airbnb & 74 & They're G-r-r-reat! - Frosted Flakes \\
25 & The Power of Dreams - Honda & 75 & Think Outside the Bun - Taco Bell \\
26 & Open Happiness - Coca-Cola & 76 & Moving Forward - Toyota \\
27 & Better Ingredients. Better Pizza. - Papa John's & 77 & Solutions for a Small Planet - IBM \\
28 & When You Care Enough to Send the Very Best - Hallmark & 78 & Breakfast of Champions - Wheaties \\
29 & Melts in Your Mouth, Not in Your Hands - M\&Ms & 79 & Eat Fresh - Subway \\
30 & Designed for Driving Pleasure - BMW & 80 & Snap! Crackle! Pop! - Rice Krispies \\
31 & You're in Good Hands - Allstate & 81 & Every Little Helps - Tesco \\
32 & Challenge Everything - Electronic Arts & 82 & The King of Beers - Budweiser \\
33 & Play. Laugh. Grow. - Fisher-Price & 83 & Keep Going - Energizer \\
34 & Easy, Breezy, Beautiful - CoverGirl & 84 & The Quicker Picker Upper - Bounty \\
35 & The Best Part of Waking Up is Folgers in Your Cup - Folgers & 85 & Fly the Friendly Skies - United Airlines \\
36 & It Takes a Licking and Keeps on Ticking - Timex & 86 & Have More Fun - Nintendo \\
37 & America Runs on Dunkin - Dunkin' Donuts & 87 & We Try Harder - Avis \\
38 & Life's Good - LG & 88 & Make the Most of Now - Vodafone \\
39 & The Best a Man Can Get - Gillette & 89 & There's No Substitute for Quality - Toyota \\
40 & Once You Pop, You Can't Stop - Pringles & 90 & Don't Just Book It, Thomas Cook It - Thomas Cook \\
41 & Don't Crack Under Pressure - Tag Heuer & 91 & Stronger than Dirt - Ajax \\
42 & Share Moments. Share Life. - Kodak & 92 & That Was Easy - Staples \\
43 & Taste the Rainbow - Skittles & 93 & Taste So Good, Cats Ask for It by Name - Meow Mix \\
44 & The World's Local Bank - HSBC & 94 & When It Rains, It Pours - Morton Salt \\
45 & Look Sharp, Feel Sharp - Gillette & 95 & Driven by Passion - Fiat \\
46 & Share the Fantasy - Chanel No. 5 & 96 & Do More - American Express \\
47 & Connecting People - Nokia & 97 & Let Your Fingers Do the Walking - Yellow Pages \\
48 & What's in Your Wallet? - Capital One & 98 & Be More Dog - O2 \\
49 & Is It In You? - Gatorade & 99 & Because Change Happens - Zurich Insurance \\
50 & I Am What I Am - Reebok & 100 & Hello Moto - Motorola \\
\hline
\end{tabular}
}
\caption{\textbf{Ranking advertising slogans with the biased BT model.} $\epsilon = 0.075$ (LLM: ChatGPT).}
\end{table} 

\begin{figure}[H]
  \centering
  \includegraphics[width=6.5in]{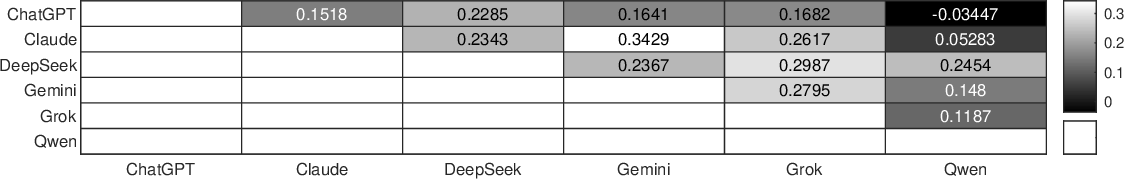}
  \caption{\textbf{Correlations of the ranking of slogans with different LLMs.} Measured by Spearman's $\rho$.}
  \label{figS2}
\end{figure}

\end{document}